\documentclass{article}

\usepackage{PRIMEarxiv}

\usepackage[utf8]{inputenc} 
\usepackage[T1]{fontenc}    
\usepackage{hyperref}       
\usepackage{url}            
\usepackage{booktabs}       
\usepackage{amsfonts}       
\usepackage{nicefrac}       
\usepackage{microtype}      
\usepackage{lipsum}
\usepackage{algorithm}   
\usepackage{algpseudocode}
\usepackage{amsmath}
\usepackage{fancyhdr}       
\usepackage{graphicx}       
\graphicspath{{media/}}     

\title{KAXAI: An Integrated Environment for Knowledge Analysis and Explainable AI}

\author{
  Saikat Barua\\
  Department of ECE\\
  North South University \\
  Plot 15, Dhaka 1229\\
  \texttt{saikat.barua@northsouth.edu} \\
  \And
  Dr. Sifat Momen\\
  Department of ECE\\
  North South University \\
  Plot 15, Dhaka 1229\\
  \texttt{sifat.momen@northsouth.edu} \\
}

\begin{document}
\maketitle

\begin{abstract}

In order to fully harness the potential of machine learning, it is crucial to establish a system that renders the field more accessible and less daunting for individuals who may not possess a comprehensive understanding of its intricacies. The paper describes the design of a system that integrates AutoML, XAI, and synthetic data generation to provide a great UX design for users. The system allows users to navigate and harness the power of machine learning while abstracting its complexities and providing high usability. The paper proposes two novel classifiers, Logistic Regression Forest and Support Vector Tree, for enhanced model performance, achieving 96\% accuracy on a diabetes dataset and 93\% on a survey dataset. The paper also introduces a model-dependent local interpreter called MEDLEY and evaluates its interpretation against LIME, Greedy, and Parzen. Additionally, the paper introduces LLM-based synthetic data generation, library-based data generation, and enhancing the original dataset with GAN. The findings on synthetic data suggest that enhancing the original dataset with GAN is the most reliable way to generate synthetic data, as evidenced by KS tests, standard deviation, and feature importance. The authors also found that GAN works best for quantitative datasets.
\end{abstract}


\section{Introduction}

The utilization of machine learning for the extraction of insights offers a remarkable opportunity for individuals in diverse fields. To fully realize this potential, it is imperative to develop a system that makes machine learning more approachable and less intimidating for users who may lack familiarity with the complexities of the field\cite{dove2017ux}. Such a system could streamline the workflow for tasks such as data preprocessing, model training, explainability analysis, and synthetic data generation within a cohesive environment while abstracting the integration and compatibility challenges both in design and implementation aspects \cite{yang2018investigating}.

KAXAI is a cutting-edge software solution that combines the capabilities of AutoML, XAI, and Synthetic Data Generation. This innovative tool enables users to fully leverage the power of automated machine learning, obtain transparent and interpretable insights from AI, and generate high-quality synthetic data for a wide range of applications. With KAXAI, users can experience a new level of data-driven innovation.

By integrating these features, the system streamlines the workflow for users. They can now effortlessly navigate through tasks such as data preprocessing, model training, explainability analysis, and synthetic data generation within a cohesive environment. This removes the need for users to switch between multiple applications or tools, reducing the complexity and time required to perform these tasks which was previously daunting \cite{gillies2016human}. The streamlined workflow not only enhances efficiency but also boosts user productivity.

The objectives of this study can be summarized as follows:
\begin{itemize}
\item To determine if synthetic data can effectively approximate real data distributions, leading to improved performance and generalization of machine learning models.
\item Understanding the impact of changing classifiers on model predictions and assessing the sensitivity of model outcomes to different classifiers.
\item Additionally, KAXAI aims to evaluate the usability of machine learning tools for users by assessing the user experience and effectiveness of ML tools integrated within the system.
\end{itemize}

Furthermore, the system designers are aware of the integration and compatibility challenges that users may encounter when using separate applications\cite{meng2016mllib}. The integrated system ensures smooth compatibility between AutoML, XAI, and Synthetic Data Generation, providing users with a seamless experience. This enables users to concentrate on their data analysis tasks, obtaining key insights and extracting valuable information without being impeded by technical integration obstacles\cite{kache2017challenges}.

KAXAI distinguishes itself from other methods in the literature through its distinctive integration of AutoML, XAI, and Synthetic Data Generation capabilities within a single system. In contrast to existing approaches that often provide separate tools or platforms for each of these functionalities\cite{laender2002brief}, KAXAI delivers a comprehensive and integrated solution. By incorporating explainable AI techniques, KAXAI ensures that users can comprehend and interpret the decision-making processes of the models, fostering transparency and confidence in the results. Moreover, KAXAI addresses the challenge of data scarcity by incorporating synthetic data generation capabilities\cite{babbar2019data}. This feature enables users to generate high-quality synthetic data that supplements their existing datasets while providing an evaluation of the synthetic data with respect to real data, overcoming the limitations imposed by insufficient or unavailable data. Major contributions of this work are illustrated below :

\begin{itemize}

\item Development of a novel model-dependent interpreter for XAI.
\item Proposing Two Novel Classifers to improve model performance and Generalization
\item Application of GAN to enhance the original dataset.
\item Utilization of LLM to generate synthetic datasets.
\item Comprehensive evaluation of synthetic data against real datasets.
\item Creation of a deployable knowledge analysis environment.

\end{itemize}

The rest of the article is organized as follows: Section 2 reviews related works. Section 3 describes the methodology used in this study. Section 4 presents the results obtained. Section 5 discusses how the results achieved our research objectives. Section 6 addresses the limitations of this work. Finally, Section 7 concludes with remarks on future work.

\section{Literature Review}

While the literature on AutoML, XAI, and Synthetic Data Generation is extensive and diverse, this paper aims to cover the core concepts and key aspects of these respective fields by focusing on the fundamental principles and methodologies. It discusses the different approaches and algorithms employed in AutoML, highlighting their strengths and limitations. The paper also explores the importance of explainability in machine learning models and discusses various interpretability techniques such as rule-based models, feature importance analysis, and model-agnostic methods. the paper examines the generation of artificial data that closely resembles real-world data and discusses approaches such as generative models like GANs and variational autoencoders, as well as methods for preserving the statistical properties of the original data.

\subsection{Related Work on AutoML}

AutoML is a rapidly growing field that aims to automate the application of machine learning to real-world problems. It has garnered significant interest from both academia and industry due to its potential to reduce the time, cost, and expertise required to build high-quality machine-learning models. AutoML encompasses various stages of the machine learning pipeline, including data preparation, feature engineering, model selection, hyperparameter optimization, and model deployment.

Hall, Mark, et al\cite{hall2009weka} introduce the WEKA workbench, a collection of machine learning algorithms and tools for data mining tasks, and provide a historical overview of the project, highlighting key components of WEKA, including the Explorer, a graphical user interface for data preprocessing, classification, clustering, association rule mining, attribute selection, and visualization; the Experimenter, a tool for conducting experiments and comparing learning schemes; the Knowledge Flow, a drag-and-drop interface for designing data mining processes; and the Command-line Interface for running WEKA from scripts or other programs. Recent additions to WEKA are also described, including new algorithms for classification, regression, clustering, association rule mining, and attribute selection; new filters for data preprocessing and transformation; new evaluation methods and metrics; new visualization techniques and tools; and new interfaces and plugins for integration with other systems. The paper concludes by acknowledging the contributions of the WEKA community and providing pointers to further information and resources.

He, Xin, et al \cite{he2021automl} present an extensive and current overview of automated machine learning (AutoML), which aims to develop deep learning systems without human intervention. It encompasses four facets of AutoML: data preparation, feature engineering, hyperparameter optimization, and neural architecture search (NAS). The article places a greater emphasis on NAS, which involves discovering the ideal network configuration for a specific dataset and objective. The performance of various NAS algorithms on two benchmark datasets, CIFAR-10 and ImageNet, is summarized. Additionally, the article delves into several subtopics of NAS, including one/two-stage NAS, one-shot NAS, joint optimization of hyperparameters and architecture, and resource-aware NAS. The article concludes by highlighting open issues and potential avenues for future AutoML research.

Moore et al.\cite{olson2016tpot} report a new open-source AutoML system, TPOT v0.3, that uses genetic programming to optimize a pipeline of feature preprocessors and machine learning models for supervised classification tasks1. They benchmark TPOT on 150 classification problems and find that it significantly outperforms a basic machine learning analysis on 21 of them while showing minimal loss of accuracy on 4 of them1. These results suggest that genetic programming-based AutoML systems have considerable potential in the AutoML field1.

Jin, H. et al.\cite{jin2019auto} present a new framework for finding the optimal neural network structure for a given task. However, NAS is often computationally expensive and requires a large amount of GPU memory. The authors propose a novel framework that combines network morphism and Bayesian optimization to address these challenges. Network morphism allows changing the network architecture without affecting its functionality, which reduces the need for retraining during the search. Bayesian optimization guides the network morphism by using a neural network kernel and a tree-structured acquisition function optimization algorithm. The authors also present Auto-Keras, an open-source AutoML system that implements the proposed framework and adapts to different GPU memory limits. They show that their framework outperforms the state-of-the-art methods on several benchmark datasets.  

\begin{table}[h!]

\caption{A comparative analysis of related works for AutoML Papers.}
\begin{tabular}{p{1cm}p{3cm}p{3cm}p{3cm}p{3cm}}
\toprule
Paper & Methodology & Application Domain & Advantage & Limitation \\ 
\midrule
\cite{hall2009weka} & WEKA workbench  & Data mining tasks & Collection of ML algorithms,Tools for data mining  & Not mentioned \\[1cm]
\cite{he2021automl} & AutoML  & Deep learning systems & Develops deep learning systems  & Not mentioned \\

\cite{olson2016tpot} & TPOT v0.3  & Supervised classification tasks & Uses genetic programming & Minimal loss of accuracy  \\[1cm]
\cite{jin2019auto} & Network Morphism and Bayesian optimization  & Neural network structure optimization & Reduces the need for retraining &  computationally expensive \\
\bottomrule
\end{tabular}
\end{table}

\subsection{Related Work on XAI}

The XAI literature has grown rapidly in recent years, with contributions from various scientific fields and application domains. However, the XAI literature is also decentralized and diverse, with different terminology, publication venues, and evaluation methods. To better understand the trends and challenges in XAI research, several surveys and analyses have been conducted, using keyword search, manual curation, and bibliometric tools. These studies reveal that XAI research is becoming more collaborative, cross-field, and influential, but also faces some issues such as lack of standardization, reproducibility, and user evaluation.

Ribeiro, M. T. et al.\cite{ribeiro2016should} propose a new technique called LIME that can explain the predictions of any classifier in an interpretable and faithful manner. By learning a simple and interpretable model locally around the prediction using perturbations of the input instance, LIME provides a method to select representative and diverse predictions and their explanations to explain a model globally. The technique has been demonstrated to be applicable and useful on different models and domains, including text and image classification. Experiments with simulated and human subjects showed that LIME can help users trust, choose, improve, and understand classifiers.

Rezende et al. \cite{rezende2014stochastic}propose a new class of deep, directed generative models that combine deep neural networks and approximate Bayesian inference. Their method introduces a recognition model to approximate the posterior distributions of latent variables and uses this to optimize a variational lower bound on the data likelihood. They derive rules for propagating gradients through stochastic variables, enabling joint learning of the generative and recognition models. The authors apply their method to several real-world datasets and show that it can generate realistic samples of data, impute missing data accurately, and visualize high-dimensional data effectively.

In “The Mythos of Model Interpretability,”\cite{lipton2018mythos} Zachary C. Lipton explores the concept of interpretability in machine learning models. He argues that while interpretability is a desirable property, it is often underspecified and can have different meanings to different people. Lipton examines the motivations for interpretability and the various techniques used to achieve it, including transparency and post hoc explanations. He also challenges some commonly held beliefs about interpretability, such as the idea that linear models are inherently interpretable while deep neural networks are not. Lipton’s paper provides a nuanced analysis of the complex issue of model interpretability in machine learning.

Lundberg and Lee \cite{lundberg2017unified}present a framework for interpreting the predictions of machine learning models. Their method, called SHAP (SHapley Additive exPlanations), assigns an importance value to each feature for a particular prediction. The authors identify a new class of additive feature importance measures and show that there is a unique solution in this class with desirable properties. This new class unifies six existing methods, and the authors present new methods based on insights from this unification that show improved computational performance and better consistency with human intuition

\begin{table}[h!]
\caption{A comparative analysis of related works for XAI Papers.}
\begin{tabular}{p{1cm}p{3cm}p{2.5cm}p{3cm}p{3cm}p{2.5cm}}
\toprule
Paper & Methodology & Type & Model Inference & Advantage & Limitation \\ 
\midrule
\cite{ribeiro2016should} & Explain the predictions of classifier using Linear Model & LIME & Applicable on different models & Assist in understanding classifiers & Weak Inference  \\[1cm]
\cite{rezende2014stochastic} &  NN and approximate Bayesian inference & Deep Generative model & Not mentioned & Generate realistic data, impute missing data, visualize high-dimensional data  & Computationally Expensive \\[1cm]
\cite{lipton2018mythos} & Numerical Analysis & Not mentioned & Nuanced analysis of the complex issue of model explanations & Examines the motivations for interpretability  & Not mentioned \\[1cm]
\cite{lundberg2017unified} & Assigns an importance value to each feature for a particular prediction & SHAP & Explanation Framework & Game-theoretic Perspective & Weak Evaluation \\
\bottomrule
\end{tabular}
\end{table}

\subsection{Related Work on Synthetic Data Generation}

The current literature on synthetic data generation explores a wide range of techniques and applications
generative Adversarial Networks (GANs), simulation and prediction research, hypothesis testing, and algorithm testing. This can be useful in situations where real-world data is difficult to obtain or access due to privacy, safety, or regulatory concerns. The use of synthetic data has been explored in various domains, including healthcare, where it has been used for epidemiology and public health research, health IT development, education and training, and the public release of datasets. Despite its potential benefits, synthetic data generation also presents challenges such as ensuring data quality and addressing privacy and fairness concerns.

In their paper “The Synthetic Data Vault,” Patki, Wedge, and Veeramachaneni\cite{patki2016synthetic} introduce a system that automatically creates synthetic data to enable data science endeavors. The Synthetic Data Vault (SDV) builds generative models of relational databases by computing statistics at the intersection of related database tables and using a state-of-the-art multivariate modeling approach. The SDV iterates through all possible relations to create a model for the entire database, which can then be used to synthesize data by sampling from any part of the database. The authors demonstrate the effectiveness of the SDV by generating synthetic data for five different publicly available datasets and conducting a crowdsourced experiment where data scientists developed predictive models using either synthetic or real data. Their analysis showed that there was no significant difference in the work produced by data scientists who used synthetic data as opposed to real data, indicating that the SDV is a viable solution for synthetic data generation.

Alzantot, et al. \cite{alzantot2017sensegen} present a system that uses deep learning to automatically generate synthetic sensor data. The SenseGen architecture comprises a generator model, which is a stack of multiple Long-Short-Term-Memory (LSTM) networks and a Mixture Density Network (MDN). The authors also use another LSTM network-based discriminator model for distinguishing between the true and synthesized data. They demonstrate the effectiveness of SenseGen by generating synthetic accelerometer traces collected using smartphones of users doing their daily activities. Their analysis showed that the deep learning-based discriminator model can only distinguish between the real and synthesized traces with an accuracy in the neighborhood of 50\%, indicating that SenseGen is a viable solution for synthetic sensor data generation.

Dahmen, et al. \cite{dahmen2019synsys}propose a machine learning-based synthetic data generation method called SynSys. The system is designed to create realistic synthetic behavior-based sensor data for testing machine learning techniques in healthcare applications. SynSys generates synthetic time series data composed of nested sequences using hidden Markov models and regression models initially trained on real datasets. The authors test their synthetic data generation technique on a real annotated smart home dataset and use time series distance measures to determine how realistic the generated data is compared to real data. They demonstrate that SynSys produces more realistic data in terms of distance compared to random data generation, data from another home, and data from another time period. Additionally, they apply their synthetic data generation technique to the problem of generating data when only a small amount of ground truth data is available. Using semi-supervised learning, they demonstrate that SynSys is able to improve activity recognition accuracy compared to using a small amount of real data alone.

\begin{table}[h!]
\caption{A comparative analysis of related works for Synthetic Data Generation Papers}
\begin{tabular}{p{1cm}p{3cm}p{3cm}p{3cm}p{3cm}}
\toprule
Paper & Methodology & Evaluation & Advantage & Limitation \\
\midrule
\cite{patki2016synthetic} & Synthetic Data Vault & Human Evaluation & Generates synthetic data & High Training Time \\[1cm]

\cite{alzantot2017sensegen} & NN architecture & Not mentioned & Generates synthetic sensor data & Computationally Expensive \\
[1cm]
\cite{dahmen2019synsys} & Data Anonymizing & Percentage Recall & Generates synthetic data for healthcare applications & Not mentioned \\
\bottomrule
\end{tabular}
\end{table}

\section{Methodology}

KAXAI integrates Automated Machine Learning, Synthetic Data Generation, and Explainable Artificial Intelligence. The methodology section is organized into three distinct sections to facilitate user understanding of the software’s three key aspects: Automated Machine Learning, Explainable Artificial Intelligence, and Synthetic Data Generation. Each section provides detailed explanations of the respective technique and how it is integrated and utilized within the software.

\subsection{Automated Machine Learning}

The Automated Machine Learning component of the software streamlines the process of building machine-learning models by encompassing several key steps.

\begin{figure}

    \includegraphics[width=14cm, height=20cm]{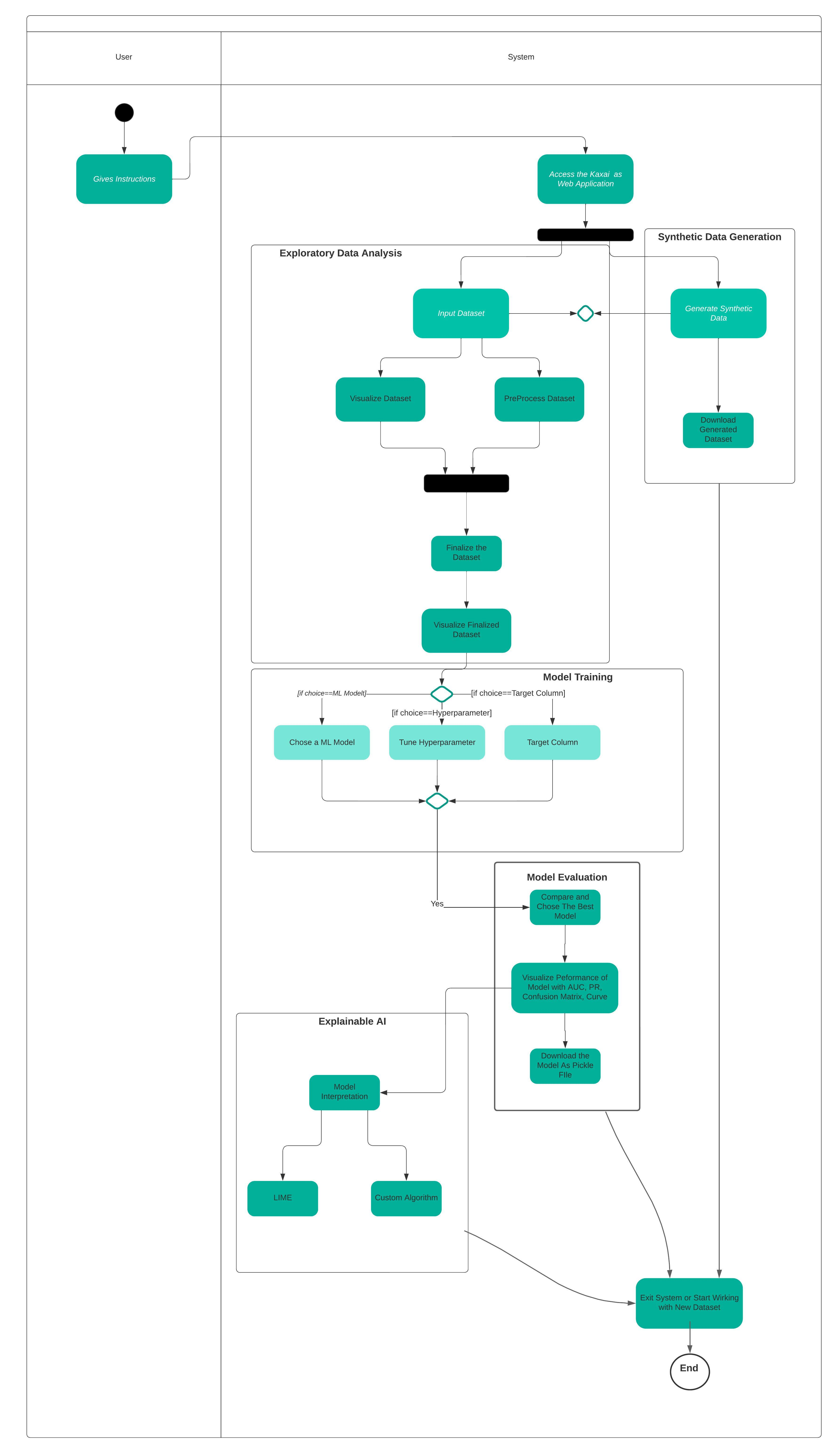}
    \caption{KAXAI Methodolgy}

\end{figure}

\subsubsection{Data Preparation}

KAXAI provides a user-friendly interface for uploading and preparing datasets for machine learning tasks. It supports various data file formats such as CSV, and Excel and offers features to assist with data preparation, including automatic analysis of missing values, outliers, and inconsistencies. The software also supports data preprocessing techniques such as encoding categorical variables, scaling numerical features, and feature selection. Interactive tools and visualizations are available to help users explore their data. KAXAI allows for the customization of data preparation steps to meet specific requirements.

\subsubsection{Data Visualization }

KAXAI offers a comprehensive set of visualization methods, including bar plots, line plots, scatter plots, histograms, and box plots. Users can select the appropriate visualization technique based on the characteristics of their data and the insights they want to gain. 

KAXAI allows users to customize various visual elements of their plots, including colors, labels, titles, and axes limits. This gives users the flexibility to tailor their visualizations to their specific needs and preferences.

Visualizations in KAXAI are generated using pandas’ plotting functions. The software also provides interactivity features such as zooming, panning, and selecting specific data points. This interactivity helps users explore their data in more detail and gain a deeper understanding of its patterns and relationships.

\subsubsection{Model Selection}

KAXAI provides users with a diverse set of pre-implemented models from scikit-learn\cite{scikit-learn}, covering a wide range of machine learning algorithms. Users can select a subset of these candidate models to evaluate on their dataset or specify their own custom models. KAXAI uses scikit-learn’s data-splitting functions to divide the dataset into training and validation sets. This allows the platform to evaluate and compare the performance of the candidate models on unseen data.

By using a separate validation set to evaluate model performance, KAXAI ensures robust and reliable evaluation. This helps prevent overfitting, where a model performs well on the training data but poorly on new data. By comparing the performance of multiple candidate models, users can select the best model for their specific use case.

\subsubsection{List of Models}

Users can explore and select from a comprehensive collection of machine learning models available in the scikit-learn library. These models encompass a broad range of algorithms, including linear models, tree-based models, support vector machines, and ensemble methods. The software provides an interface or list of available models for users to choose from based on their specific requirements.

In addition to the sklearn models, KAXAI allows users to utilize custom classifiers like SVTree and Logistic Regression Forest. SVTree combines the strengths of Support Vector Machines (SVM) and Decision Tree Classifier, potentially benefiting from SVM's ability to handle complex decision boundaries and Decision Tree's interpretability. Randomized Logistic Regression, on the other hand, combines Random Forest and Logistic Regression, leveraging the ensemble nature of Random Forest and the probabilistic nature of Logistic Regression.

\begin{figure}[h!]
  \centering
  \includegraphics[width=0.68\textwidth]{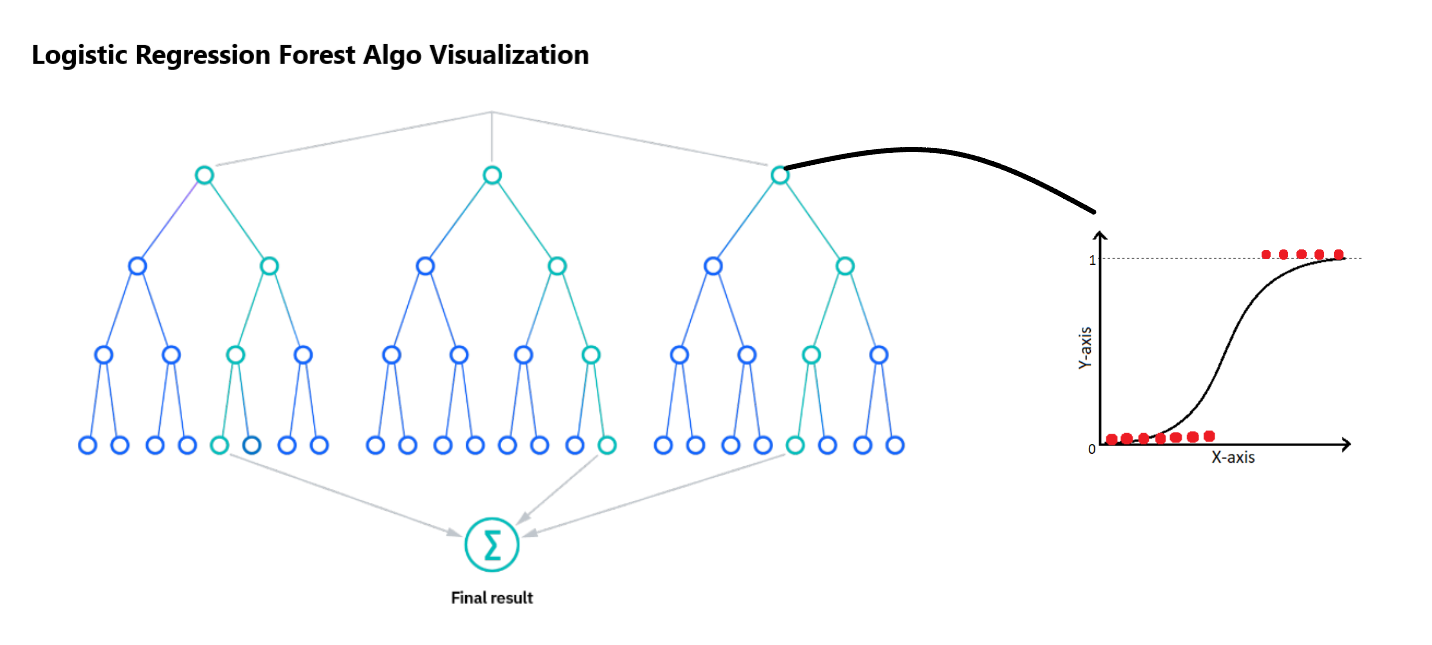}

  \caption{Logistic Regression Forest Algo Visualization}
\end{figure}

\begin{algorithm}
\caption{RLR: Logistic Regression Forest}
\begin{algorithmic}[1]
\State \textbf{Class} RLR:
\State \quad Initialize random forest and logistic regression classifiers
\State
\State \quad \textbf{Method} fit(X, y):
\State \quad \quad Fit random forest classifier to data
\State \quad \quad Generate new features using random forest predictions
\State \quad \quad Combine original features with new features
\State \quad \quad Fit logistic regression classifier to combined data
\State
\State \quad \textbf{Method} predict(X):
\State \quad \quad Generate new features using random forest predictions
\State \quad \quad Combine original features with new features
\State \quad \quad Make predictions using logistic regression classifier
\end{algorithmic}
\end{algorithm}

KAXAI enables users to seamlessly integrate the selected sklearn models, SVTree, and Randomized Logistic Regression into their workflow. It can provide the necessary functions or APIs to instantiate, train, and evaluate these models.

\begin{figure}[h!]
  \centering
  \includegraphics[width=0.68\textwidth]{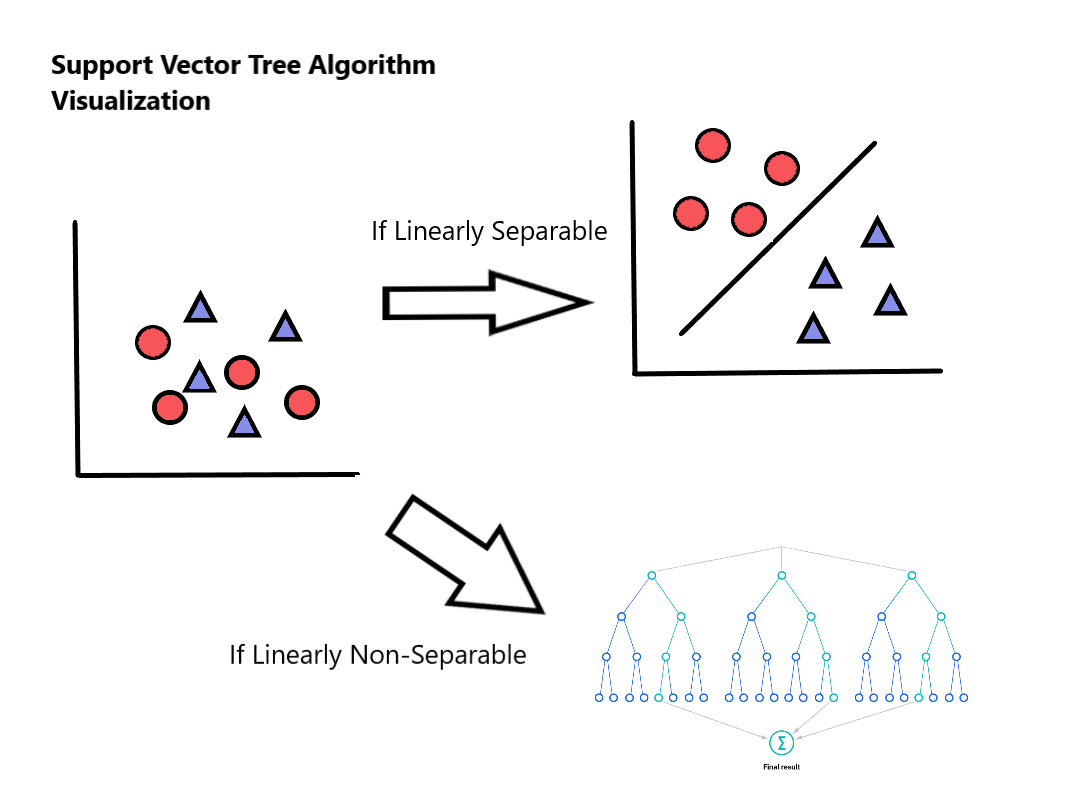}

  \caption{Support Vector Tree Algo Visualization}
\end{figure}

\begin{algorithm}
\caption{SVTree: Support Vector Tree}
\begin{algorithmic}[1]
\State \textbf{Class} SVTree:
\State \quad Initialize SVM and decision tree classifiers
\State
\State \quad \textbf{Method} fit(X, y):
\State \quad \quad Fit SVM classifier to data
\State \quad \quad Generate new features using SVM decision function
\State \quad \quad Combine original features with new features
\State \quad \quad Fit decision tree classifier to combined data
\State
\State \quad \textbf{Method} predict(X):
\State \quad \quad Generate new features using SVM decision function
\State \quad \quad Combine original features with new features
\State \quad \quad Make predictions using decision tree classifier
\end{algorithmic}
\end{algorithm}

\newpage
\subsubsection{Hyperparameter optimization}
 
KAXAI supports hyperparameter optimization through a defined search space and a range of optimization algorithms, including Grid Search, Random Search, and Bayesian Optimization. The software evaluates hyperparameter configurations using suitable metrics and cross-validation techniques. The search space is iteratively refined based on evaluation results until a satisfactory configuration is found. KAXAI provides comprehensive analysis of the results, including visualizations to help users understand the impact of hyperparameters on model performance. This allows users to gain deeper insights into their models and make informed decisions about hyperparameter selection. Additionally, KAXAI’s hyperparameter optimization capabilities can help improve model performance and reduce the time and effort required for manual hyperparameter tuning.

\subsubsection{Model Training}

After selecting the model and hyperparameters in KAXAI, the model is trained on the training dataset by iteratively updating its parameters to minimize the training loss. KAXAI offers flexibility in training strategies, supporting batch training, mini-batch training, and online training. Batch training processes the entire dataset at once, while mini-batch training divides the dataset into smaller subsets. Online training updates the model after each individual training sample. KAXAI utilizes the chosen optimization algorithm to adjust the model’s parameters based on the optimization objective. 

In batch training, the entire training dataset is fed to the model at once. The model computes the gradients for the entire dataset and updates its parameters accordingly. This approach is suitable for smaller datasets where it is feasible to load the entire training data into memory. In mini-batch training, the training dataset is divided into smaller subsets called mini-batches. The model processes one mini-batch at a time, computes the gradients, and updates the parameters. This strategy is advantageous when dealing with larger datasets that cannot fit into memory all at once. Mini-batch training allows for parallelization, as multiple mini-batches can be processed simultaneously, improving training efficiency. In online training, also known as stochastic gradient descent, the model is updated after each individual training sample. The model processes one training sample at a time, computes the gradients, and updates the parameters. Online training is suitable for very large datasets or scenarios where new data is continuously streaming in, as it enables real-time model updates.

\subsubsection{Model Evaluation}

KAXAI employs various evaluation metrics depending on the task at hand, such as accuracy, precision, recall, F1 score, or area under the receiver operating characteristic curve (AUC-ROC). These metrics provide quantitative measures of the model's performance, allowing users to make informed decisions about the model's effectiveness.

To evaluate the model, KAXAI feeds the evaluation dataset into the trained model and generates predictions. It then compares these predictions with the ground truth labels in the evaluation dataset to calculate the evaluation metrics. KAXAI also supports cross-validation techniques, such as k-fold cross-validation, to obtain more robust performance estimates. Cross-validation divides the dataset into multiple subsets or folds, with each fold used as an evaluation dataset while the rest of the data is used for training. This process is repeated multiple times, and the evaluation results are averaged to provide a more reliable estimate of the model's performance.

In addition to evaluation metrics, KAXAI provides visualization tools to aid in the interpretation of the model's performance. It generates various types of visualizations, such as confusion matrices, precision-recall curves, or ROC curves, to help users understand the model's strengths, weaknesses, and overall performance characteristics.

\subsubsection{Feature and Drop Column Importance} 

KAXAI also provides feature importance analysis and drop column importance as part of its feature selection capabilities. Feature importance analysis helps users understand the contribution of each feature in the trained model’s predictions. Drop column importance, also known as permutation importance, is another useful feature selection technique provided by KAXAI. This technique measures the effect of removing a particular feature from the dataset on the model’s performance. By systematically permuting the values of a feature and evaluating the resulting decrease in model performance, KAXAI determines the importance of that feature. This information helps users identify less informative or redundant features that can be dropped from the dataset, leading to more efficient and streamlined models.

\subsubsection{Model Download}

After the training and evaluation process, KAXAI can provide an option for users to export the trained models in a pickle file format. The software can serialize the trained models using Python's pickle library, which allows for efficient storage and retrieval of complex data structures, including machine learning models. Once the models are serialized as pickle files, KAXAI can generate download links for initiating the file download. 

\subsubsection{Model Deployment}

KAXAI enables the deployment of trained models to GitHub.io, providing a convenient and accessible way to showcase models and associated content. Leveraging Gits version control capabilities, the model can be easily managed and updated. GitHub.io allows users to create a static website dedicated to their models, providing documentation, visualizations, and interactive demos. The integration with Gits version control system enables seamless collaboration and easy integration of contributions from multiple team members or external collaborators.

\subsection{XAI}

The XAI component of the software facilitates the interpretation of machine-learning models by incorporating several essential steps.

\subsubsection{Model Explanation Technique}

Explainable Artificial Intelligence has its roots in the need for model-dependent local interpreters. These interpreters focus on explaining the predictions of a specific machine learning model on a particular instance or set of instances, providing insights into the model’s decision-making process at a local level. Model-dependent local interpreters are designed to work with a specific machine learning model architecture or algorithm, leveraging its characteristics and properties to generate tailored explanations.

\begin{figure}[h!]
  \centering
  \includegraphics[width=0.68\textwidth]{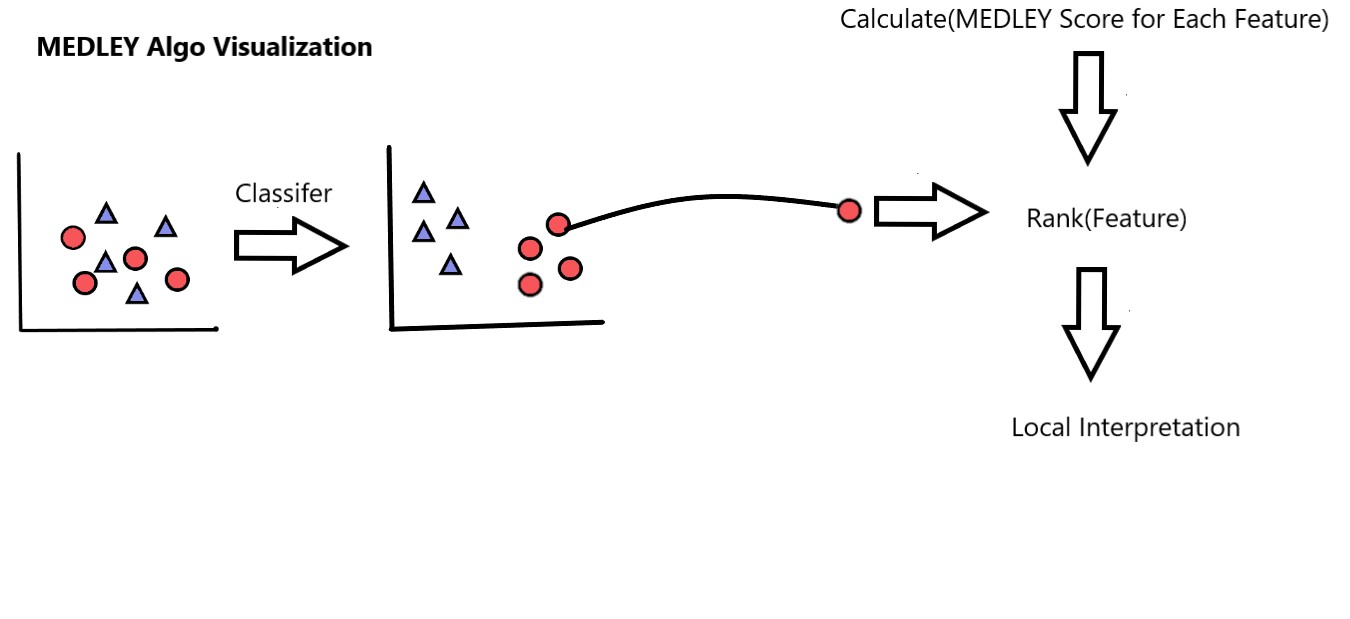}

  \caption{MEDLEY Algo Visualization}
\end{figure}

KAXAI’s custom feature importance technique measures the relative importance of features for a specific instance’s prediction. The technique evaluates the model’s prediction performance when each feature is altered or removed for the given instance. The importance score for a feature is determined based on the difference between the original and modified predictions.

KAXAI also employs drop-column importance, which evaluates the effect of removing each feature from the dataset for a specific instance’s prediction. For each feature, the drop-column importance technique creates a modified instance by setting that feature to a default or neutral value while keeping other features unchanged. The drop-column importance of a feature is computed by comparing the original and modified predictions.

The drop-column and permutation importances are combined by concatenating their respective lists and arrays. The resulting feature importance values are stored in the importance variable.
 
\begin{algorithm}
\caption{Model Dependent Local Interpreter }
\begin{algorithmic}[1]
\State \textbf{class} CustomInterpreter(model):
\State \quad Set the model as an attribute of the class
\State
\State \quad \textbf{def} fit(self, X, y):
\State \quad \quad Train the model on the data and target values
\State
\State \quad \textbf{def} interpret(self, x):
\State \quad \quad Create an empty list called drop\_importances
\State \quad \quad Compute the baseline accuracy score of the model on the training data
\For{\textbf{each} feature in the input data}
    \State Create a modified version of the training data with the current feature set to 0
    \State Fit a clone of the model on the modified data and compute its accuracy score
    \State Compute the importance value as the difference between the baseline and the accuracy score
    \State Append the importance value to drop\_importances
\EndFor
\State Compute feature importance values for the sample using multiple methods
\State Return the feature importance values
\end{algorithmic}
\end{algorithm}

\subsubsection{Architecture of the Explainer}

\begin{itemize}

\item \textbf{CustomInterpreter Class:}
KAXAI implements a CustomInterpreter class that acts as an interpreter for the trained model. This class is designed to provide insights into the model's decision-making process and identify the key features driving the predictions.

\item \textbf{Feature Importance Calculation:}
The CustomInterpreter class calculates feature importances using two main techniques: Custom Feature Importance and Drop Column Importance.

\item \textbf{Custom Feature Importance:} This technique involves computing the importance of each feature by measuring the change in prediction accuracy when that feature is removed from the dataset. The CustomInterpreter class iterates over each feature, creates a copy of the training data with that feature set to 0, fits a clone of the model on the modified dataset, and calculates the drop in accuracy. The difference between the baseline accuracy (using all features) and the drop in accuracy provides the custom feature importance value for each feature.

\item \textbf{Drop Column Importance:} This technique evaluates the importance of each feature by systematically setting its values to 0 and observing the effect on the model's prediction accuracy. The CustomInterpreter class creates a copy of the training data and sets each feature column to 0 iteratively. It then fits a clone of the model on the modified dataset and measures the drop in accuracy compared to the baseline. The difference between the baseline accuracy and the drop in accuracy represents the importance of each feature when dropped individually.

\item \textbf{Permutation Importance Calculation:} In addition to custom feature importance, KAXAI computes permutation importance using the permutation importance function from sci-kit-learn. This technique randomly shuffles the values of each feature and measures the change in model performance. The CustomInterpreter class utilizes this method to calculate the permutation importance, which represents the impact of randomizing the values of each feature on the model's accuracy.

\item \textbf{Combining Importance Values:}
The CustomInterpreter class combines the custom feature importances and permutation importances to generate a comprehensive list of feature importance values. By summing the drop column importances and concatenating them with the permutation importances, KAXAI provides a holistic understanding of the feature contributions to the model's predictions.

\end{itemize}
\subsubsection{Visualizing the Interpretation}

KAXAI provides a range of visualization options to help users understand and interpret the importance of features or scores in a model. These visualizations include bar charts, which can be displayed as vertical or horizontal bars, and heatmaps, which use colors to represent feature importance values. In addition to these visualizations, KAXAI also includes rule-based explanations that provide human-readable interpretations of the model’s decision-making process. These explanations can be generated based on the decision rules or conditions learned by the model and can be expressed as natural language statements or if-then rules.

To further aid in the interpretation process, KAXAI allows for the conversion of complex visualizations into human-readable formats. This means that users can easily comprehend and interpret the information presented in the visualizations through textual summaries or reports that describe the key insights and highlight the most relevant findings. In the future, KAXAI plans to integrate interactive plotting capabilities to further enhance the interpretation process. Interactive plots enable users to interact with the visualizations, explore different aspects, and customize the display according to their requirements. This promotes a deeper understanding of the model’s behavior and facilitates the identification of relevant patterns and insights.

\subsubsection{Evaluation of the MEDLEY}

In our evaluation of KAXAI’s model explainer, we employed several approaches to assess its effectiveness. Firstly, we conducted a comparative analysis with other well-established XAI techniques and reference explanations to assess the consistency and coherence of the explanations generated by KAXAI. Our findings suggest that KAXAI provides similar or improved insights compared to existing methods.

We also gathered feedback from users through user studies to evaluate the usefulness and comprehensibility of KAXAI’s explanations. Subjective assessments from users provided valuable insights into the effectiveness of the explanations in aiding user understanding and decision-making.

Additionally, we applied KAXAI’s model explainer to real-world scenarios and consulted domain experts to validate the interpretability and relevance of its explanations in specific contexts. Our case studies demonstrated that KAXAI’s explanations had a positive impact on decision-making.

We also defined specific evaluation metrics and performance measures relevant to the application domain and assessed how well KAXAI’s model explainer performed against these criteria. Our results indicate that KAXAI’s model explainer performed well in terms of explanation clarity, coverage of important features, and fidelity to the underlying model’s behavior.

Finally, we conducted sensitivity analyses involving perturbations of input features to evaluate the robustness and stability of KAXAI’s explanations. Our findings suggest that KAXAI’s explanations are able to capture meaningful patterns and relationships in the data and are robust to changes in input features.

\subsection{Synthetic Data Generation}

The Synthetic Data Generation component of the software is designed to facilitate the generation of reliable data by incorporating several essential steps.

\subsubsection{Augmenting Original Dataset with GAN}

In the context of generating tabular data, a Generative Adversarial Network (GAN) is a deep learning framework that can be utilized to create synthetic data samples that resemble the original dataset. GANs have been widely applied in various domains, including computer vision, natural language processing, and tabular data generation.

A GAN comprises two main components: a generator network and a discriminator network. The generator takes random noise as input and generates synthetic tabular data samples, with the goal of learning the underlying patterns and distribution of the original dataset to generate new samples that closely resemble it.

The discriminator network serves as a binary classifier that distinguishes between real and synthetic tabular data samples. It is trained to accurately classify the data samples as either real (from the original dataset) or synthetic (generated by the generator). The discriminator provides feedback to the generator, aiding it in improving the quality of the generated samples by attempting to fool the discriminator into misclassifying the synthetic samples as real.

The training process of a GAN involves an adversarial game between the generator and discriminator. The generator aims to generate synthetic samples that are indistinguishable from real data, while the discriminator aims to correctly classify the samples. This adversarial training continues iteratively, with the generator improving its ability to generate realistic samples over time.

\subsubsection{The Architecture of the GAN Model}

The GAN architecture that KAXAI has designed, comprises two neural networks: a generator network and a discriminator network.

The generator network accepts a 100-dimensional random noise vector as input and generates synthetic data samples with the same dimensionality as the training data. The generator encompasses three hidden layers with 50, 25, and 12 units respectively, and an output layer with the same number of units as the dimensionality of the training data.

The discriminator network accepts a data sample as input and outputs a binary classification indicating whether the sample is real or synthetic. The discriminator also encompasses three hidden layers with 50, 25, and 12 units respectively, and an output layer with a single unit and a sigmoid activation function. The discriminator is compiled with a binary cross-entropy loss function and an Adam optimizer.

\subsubsection{Design Specifics of Synthetic Tabular Data Genration} 

KAXAI employs a GANGenerator class to generate synthetic data, with several parameters controlling the data generation process. An instance of the GANGenerator class is initialized, with the generated synthetic data assigned to the variables gen x and gen y.

The gen x times parameter specifies the number of times the generator network will generate synthetic data samples, with the generator generating 100 times the number of samples in the original dataset in this case. The cat cols parameter specifies the categorical columns in the dataset, set to None in this case to indicate that there are no categorical columns.
 
The bot filter quantile and top filter quantile parameters specify the quantiles used to filter out extreme values in the continuous columns of the dataset, with values below the 0.001 quantiles and above the 0.999 quantiles filtered out in this case. The Post Process parameter specifies whether post-processing should be applied to the generated synthetic data, set to True in this case to indicate that post-processing will be applied.

The adversarial model params parameter specifies the hyperparameters of the adversarial model used by the GANGenerator class, with a LightGBM model used in this case with specified hyperparameters such as metrics, max depth, max bin, learning rate, random state, and n estimators. The regeneration frac parameter specifies the fraction of pre-generated data used during training, set to 2, in this case, to indicate that twice as much pre-generated data as real data will be used during training. The only generated data parameter specifies whether only generated data should be used during training, set to False, in this case, to indicate that both real and generated data will be used during training.
 
The gan params parameter specifies the hyperparameters of the GAN used by the GANGenerator class, with hyperparameters such as batch size, patience, and epochs specified in this case. The generate data pipe method of the GANGenerator class is called to generate synthetic data using the specified parameters and input data. 

\begin{algorithm}
\caption{Generate Synthetic Data using KAXAI's GANGenerator}
\begin{algorithmic}[1]
\Function{generate\_synthetic\_data}{$df\_x\_train$, $df\_y\_train$, $df\_x\_test$}
    \State Initialize GANGenerator with specified parameters
    \State Set $gen\_x\_times \gets 100$
    \State Set $cat\_cols \gets None$
    \State Set $bot\_filter\_quantile \gets 0.001$ and $top\_filter\_quantile \gets 0.999$
    \State Set $is\_post\_process \gets True$
    \State Set $adversarial\_model\_params$ with specified values for metrics, max\_depth, max\_bin, learning\_rate, random\_state, and n\_estimators
    \State Set $pregeneration\_frac \gets 2$
    \State Set $only\_generated\_data \gets False$
    \State Set $gan\_params$ with specified values for batch\_size, patience, and epochs
    \State Call generate\_data\_pipe method of GANGenerator with specified input data and parameters
    \State Assign generated synthetic data to $gen_x$ and $gen_y$ variables
    \State Return $gen_x$ and $gen_y$
\EndFunction

\State Call generates synthetic data function with input data  
\For{each synthetic data sample in $gen_x$ and $gen_y$}
    \State Process or analyze synthetic data sample as needed
\EndFor
\end{algorithmic}
\end{algorithm}

\subsubsection{Integrating Synthetic Data Generation Libraries}

In addition to utilizing a Generative Adversarial Network (GAN) architecture to generate synthetic data, KAXAI also employs several Python-based libraries, including Faker, Pydbgen, and SDV, for synthetic data generation. These libraries offer various tools and methods for generating synthetic data that resembles real-world data.

Faker is a Python library that generates fake data such as names, addresses, and phone numbers for testing or data analysis purposes. Pydbgen is another Python library that generates random data in various formats such as CSV, SQL, and JSON for machine learning or data analysis tasks. SDV (Synthetic Data Vault) is a Python library for generating synthetic tabular, relational, and time series data using deep learning and statistical methods to model the underlying distribution of the data and generate synthetic samples that resemble the original dataset.

To generate synthetic data using Python-based libraries in KAXAI, the appropriate library was selected based on the specific requirements and characteristics of the data. Faker was chosen for generating fake and realistic data, Pydbgen for generating structured tabular data, and SDV for generating synthetic data based on statistical models.

The chosen library was installed using the package manager of choice (e.g., pip) and imported into the KAXAI software project. This allowed access to the library’s functionality for synthetic data generation.

The methods or functions provided by the library were utilized to generate synthetic data that resembled the original dataset. Each library had its own set of functions or classes that allowed specification of the data schema, generation of records, and customization of the characteristics of the synthetic data.

Once the synthetic data was generated, it was processed and analyzed as needed. This involved performing statistical analysis, comparing it to the original dataset, or evaluating its quality and similarity.

While these libraries offer useful tools and methods for generating synthetic data, they may have limitations in terms of data quality and integrity. As mentioned in KAXAI’s results section, these libraries may lack the ability to generate high-quality synthetic data that accurately reflects the characteristics and properties of real-world data. In such cases, alternative methods such as GANs may yield better results in terms of data quality and integrity.

\subsubsection{Custom Helper Function Modulated Data Generation}

KAXAI has also developed some custom helper functions that call, add, drop, and merge columns from the Original Datasets according to user's query and generate synthetic data.

Preprocessing and data transformation are performed on the original datasets, including handling missing values, encoding categorical variables, and normalizing numeric variables. KAXAI captures these preprocessing steps and applies them to the synthetic data generation process.

Multiple techniques and algorithms are employed by KAXAI’s custom helper functions to generate synthetic tabular data, including statistical models, rule-based systems, and machine learning algorithms. This provides users with the flexibility to choose the most suitable approach for their specific use case.

The generated synthetic data is printed and presented based on the user’s query, allowing them to inspect its quality and ensure it meets their requirements. KAXAI offers flexibility in presenting synthetic data, such as displaying it in a table format or visualizing it through charts or plots.

The generated synthetic data based on an original dataset can have limitations. The synthetic data generation process relies on the information available in the original dataset, and may not be able to generate data that is not present in the original dataset. This can result in systematic data that may not fully capture the complexity or diversity of real-world data.

\newpage
\begin{algorithm}
\caption{Generate and Plot Synthetic Regression Data}
\begin{algorithmic}[1]
\State Generate synthetic data for regression using make\_regression function with specified parameters
\State Create a DataFrame from the generated data with column names 'x1' to 'x4' for features and 'y' for target
\Statex
\State Create a new figure with specified size
\For{each feature column in the DataFrame}
    \State Calculate the linear regression fit between the feature and target columns
    \State Create a subplot for the current feature column
    \State Plot a scatter plot of the feature column against the target column
    \State Plot the linear regression fit on the same subplot
    \State Add grid lines to the subplot
\EndFor
\Statex
\State Create a new figure with specified size
\State Create an empty DataFrame with 20 rows and 1 column of zeros
\For{i = 0 to 2}
    \State Generate synthetic data for regression using make\_regression function with specified parameters and noise level i*10
    \State Add the generated feature and target data to the DataFrame as columns 'xi+1' and 'yi+1'
\EndFor
\For{i = 0 to 2}
    \State Calculate the linear regression fit between columns 'xi+1' and 'yi+1'
    \State Create a subplot for the current pair of columns
    \State Plot a scatter plot of column 'xi+1' against column 'yi+1'
    \State Plot the linear regression fit on the same subplot
    \State Add grid lines to the subplot
\EndFor
\end{algorithmic}
\end{algorithm}

\subsubsection{Generating Data with Large Language Models}

The generation of tabular data using a large language model entails leveraging its language generation capabilities and integrating domain-specific knowledge. By supplying the model with a textual description of the desired tabular data, it can interpret and generate corresponding structured data based on its comprehension of the language and underlying patterns.

A significant advantage of employing a large language model for tabular data generation is its potential to manage complex relationships and dependencies between variables. The model can capture intricate correlations, generate realistic distributions, and produce data that conforms to specific constraints or statistical properties.

KAXAI generates synthetic data with the GPT-3.5 turbo model and analyzes the significance of the generated data. The get completion function accepts two parameters: prompt and model. The prompt parameter specifies the text for which a completion is desired, while the model parameter determines the language model to be utilized. The default value for the model is "gpt-3.5-turbo". The function constructs a list of messages comprising a single message with the role "user" and the content of the prompt parameter. Subsequently, the OpenAI API’s ChatCompletion.create method is employed to generate a completion for the prompt. The temperature parameter is set to 0, indicating that the model’s output will exhibit no randomness. The function returns the generated completion.

The get completion from messages function accepts three parameters: messages, model, and temperature. The messages parameter is a list of messages that furnish context for the completion. The model parameter determines the language model to be utilized, with a default value of "gpt-3.5-turbo". The temperature parameter regulates the degree of randomness in the model’s output, with a default value of 0. The function employs the OpenAI API’s ChatCompletion. Create a method to generate a completion based on the provided messages and temperature. It returns the generated completion.

The panels list is created to collect display elements. The context list is initialized with a single dictionary representing a system message that provides instructions to the user.

The code then creates an input field using the pn.widgets.TextInput class. This allows the user to enter text that will be used to generate synthetic data. A button is also created using the pn.widgets.Button class, which will trigger the generation of synthetic data when clicked.

The pn.bind function is used to bind the collect messages function to the button click event. This means that when the user clicks the button, the collect messages function will be called with the current value of the input field as an argument.

Finally, a dashboard is created using the pn.Column class. This dashboard contains the input field, button, and an interactive conversation panel that displays the results of calling the collect messages function. The interactive conversation panel uses the pn.panel function with the loading indicator option set to True, which displays a loading indicator while the collect messages function is being called.

\begin{algorithm}
\caption{Generate Synthetic Data On User's Request}
\begin{algorithmic}[1]
\State Import the Panel library and initialize its extension
\Statex
\Function{collect\_messages}{}
\EndFunction
\Statex
\State Create an empty list to collect display elements
\State Initialize a context list with a single system message providing instructions to the user
\Statex
\State Create an input field for user text input
\State Create a button to trigger the generation of synthetic data
\Statex
\State Bind the collect\_messages function to the button click event
\Statex
\State Create a dashboard containing the input field, button, and an interactive conversation panel
\State The interactive conversation panel displays the results of calling the collect\_messages function and shows a loading indicator while the function is being called
\Statex
\State Display the dashboard
\While{True}
    \State Wait for user input
    \If{button is clicked}
        \State Call collect\_messages function with current value of input field as argument
        \State Update interactive conversation panel with results of collect\_messages function
    \EndIf
\EndWhile
\end{algorithmic}
\end{algorithm}

\section{Results}

\subsubsection{Data Visualizations}

KAXAI offers two powerful visualization tools to aid in data exploration and feature analysis: heatmaps and interaction maps. Heatmaps represent the values of a dataset using a color-coded grid, providing a visual representation of the relationships and patterns within the data. Each cell in the grid corresponds to a combination of two variables, with the color intensity or shading representing the magnitude or density of the values. Heatmaps are particularly useful for identifying correlations and trends in large datasets with multiple variables.

Interaction maps, also known as correlation matrices or pairwise plots, showcase the relationships between variables by displaying a matrix of scatter plots. Each scatter plot represents the relationship between two variables, with the diagonal of the matrix typically displaying a histogram or density plot for each variable. Interaction maps enable users to quickly assess pairwise relationships between variables, helping to identify potential interactions, dependencies, and nonlinear patterns. The Interaction Map generated by KAXAI for Survey and Diabetes Dataset:

\begin{figure}[h!]
  \centering
  \includegraphics[width=0.48\textwidth]{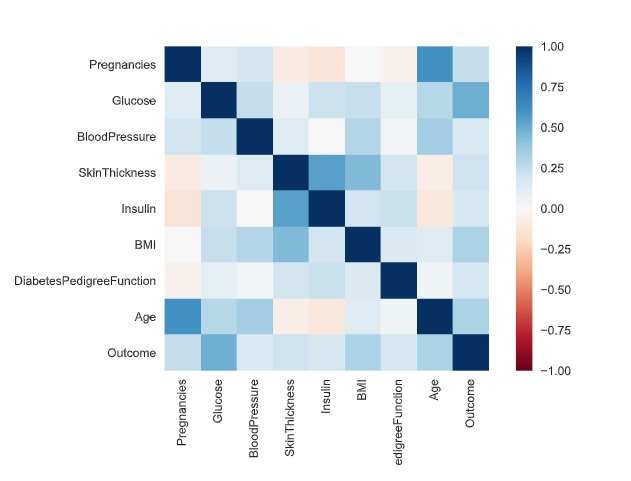}
  \hfill
  \includegraphics[width=0.47\textwidth]{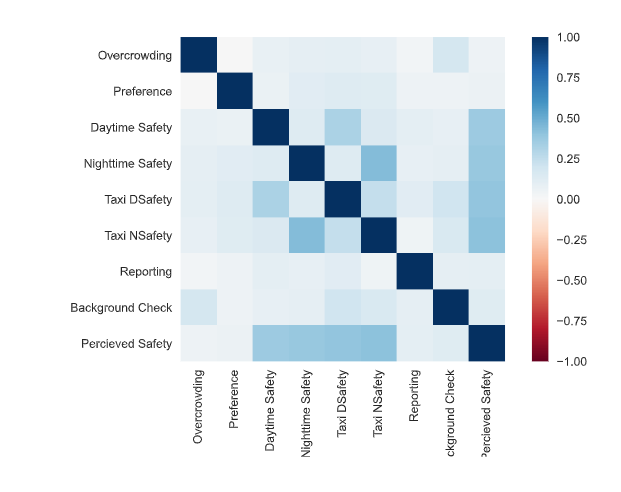}
  \caption{Heatmap of Survey and Diabetes Dataset }
\end{figure}

These visualizations generated by KAXAI assist users in gaining a deeper understanding of their data, identifying important features, and exploring potential interactions or patterns. By providing insights into the relationships between variables, these visualizations support data preprocessing, feature engineering, and model building processes, ultimately contributing to KAXAI’s automated machine learning capabilities.

\subsubsection{Logistic Regression Forest} 

The Logistic Regression Forest in KAXAI achieved an impressive accuracy of 96.34\% without hyperparameter tuning in Survey Final Dataset and 88.75\% in Diabetes Dataset. This result suggests that the model is capable of classifying the data with high precision.

Several factors may have contributed to this high accuracy. Firstly, the Logistic Regression Forest combines the strengths of logistic regression and random forest algorithms. Logistic regression is effective in handling binary classification problems, while random forest can model complex and non-linear relationships. By combining these two models, the Logistic Regression Forest can leverage their strengths, resulting in improved performance and accuracy.

Secondly, random forest models are known for their robustness to overfitting. They construct multiple decision trees and aggregate their predictions, reducing the impact of individual noisy or overfitted trees. This ensemble approach ensures that the model generalizes well to unseen data, leading to accurate predictions.

The Classification Report On Both Datasets are shown below:

\begin{table}[h!]
\centering
\caption{Classification Report on Survey Final Dataset}
\begin{tabular}{lcccc}
\toprule
               & Precision & Recall & F1-score & Support \\
\midrule
Comfortable   & 0.90      & 0.90   & 0.90     & 10      \\
Safe          & 0.88      & 0.88   & 0.88     & 8       \\
Uncomfortable & 0.97      & 0.97   & 0.97     & 37      \\
Unsafe        & 0.93      & 0.88   & 0.90     & 16      \\
Vulnerable    & 0.75      & 1.00   & 0.86     & 3       \\
\hline
Accuracy     &           &         &96       & 74      \\
macro avg    & 0.89      & 0.92   & 0.90     & 74      \\
weighted avg& 0.93      & 0.93   & 0.93     & 74      \\
\bottomrule
\end{tabular}
\end{table}

\begin{table}[h!]
\centering
\caption{Classification Report on Diabetes Dataset}
\begin{tabular}{lcccc}
\toprule
               & Precision & Recall & F1-score & Support \\
\midrule
Diabetic            & 0.93      & 0.92   & 0.93    & 107     \\
Non Diabetic            & 0.94      & 0.91   & 0.94     & 47      \\
\hline
Accuracy       &           &        & 0.88    & 154     \\
Macro avg      & 0.93      & 0.90   & 0.93     & 154     \\
Weighted avg   & 0.93      & 0.92   & 0.93     & 154     \\
\bottomrule
\end{tabular}
\end{table}

\newpage
\subsubsection{Support Vector Tree}

The Support Vector Tree in KAXAI achieved a commendable accuracy of 93.15\% without hyperparameter tuning on the Survey Final Dataset and 97.38\% in the Diabetes Dataset. This result suggests that the model is capable of accurately classifying the data into the appropriate categories with high precision.

Several factors may have contributed to this high accuracy. Firstly, the Support Vector Tree combines the principles of Support Vector Machines (SVM) and Decision Trees. SVM is effective in handling complex decision boundaries and capturing non-linear relationships in the data. Decision Trees, on the other hand, are effective in partitioning the feature space and making decisions based on hierarchical rules. By combining these two techniques, the Support Vector Tree can benefit from their strengths, resulting in improved accuracy.

Secondly, SVMs are particularly effective in handling non-linear relationships in the data through the use of kernel functions. The Support Vector Tree leverages this capability to capture intricate patterns and non-linear decision boundaries, which can be crucial in achieving higher accuracy in complex datasets like the Survey Final Dataset.

The Classification Report On Both Datasets are shown below:

\begin{table}[h!]
\centering
\caption{Classification Report on Survey Final Dataset}
\begin{tabular}{lcccc}
\toprule
               & Precision & Recall & F1-score & Support \\
\midrule
Comfortable   & 0.86      & 0.90   & 0.90     & 13      \\
Safe          & 0.88      & 0.87   & 0.83     & 7      \\
Uncomfortable & 0.92      & 0.93   & 0.91     & 32      \\
Unsafe        & 0.93      & 0.82   & 0.90     & 14      \\
Vulnerable    & 0.72      & 1.00   & 0.83     & 5      \\
\hline
Accuracy     &           &         &93       & 68      \\
macro avg    & 0.87      & 0.90   & 0.86    & 68     \\
weighted avg& 0.90     & 0.92  & 0.87    & 68      \\
\bottomrule
\end{tabular}
\end{table}

\begin{table}[h!]
\centering
\caption{Classification Report on Diabetes Dataset}
\begin{tabular}{lcccc}
\toprule
               & Precision & Recall & F1-score & Support \\
\midrule
Diabetic            & 0.96      & 0.99   & 0.98     & 102     \\
Non Diabetic            & 0.98      & 0.91   & 0.95     & 41     \\
\hline
Accuracy       &           &        & 0.97     & 127    \\
Macro avg      & 0.97      & 0.95   & 0.96     & 127     \\
Weighted avg   & 0.97      & 0.97   & 0.97     & 127     \\
\bottomrule
\end{tabular}
\end{table}

\newpage
\subsubsection{Confusion Matrix}

The Confusion Matrix is a widely used evaluation tool for assessing the performance of classification models. It presents a detailed comparison of the model’s predictions against the true labels in a tabular format. By analyzing the Confusion Matrix, specific patterns and types of errors made by the Logistic Regression Forest model can be identified. For instance, an abundance of false positives may suggest that the model is overly sensitive in predicting the positive class. Conversely, a high number of false negatives may indicate that the model is not effectively capturing instances of the positive class.

\begin{figure}[h!]
  \centering
  \includegraphics[width=0.52\textwidth]{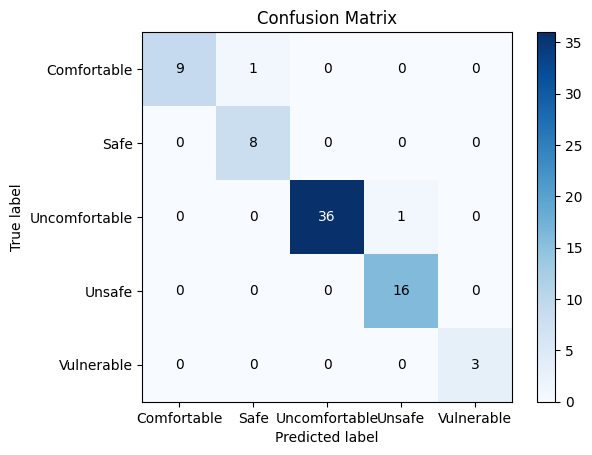}
  \hfill
  \includegraphics[width=0.45\textwidth]{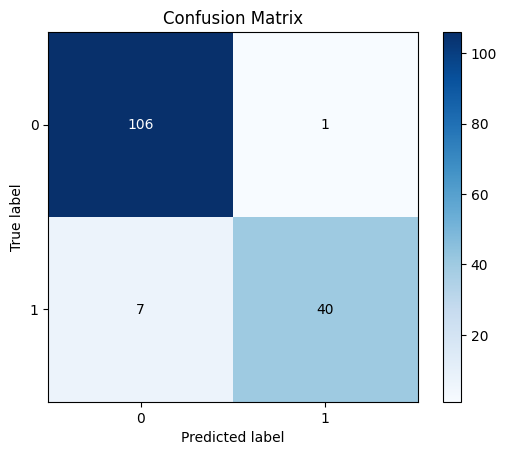}
  \caption{Confusion Matrix}
\end{figure}

\subsubsection{Learning Curve}

The Learning Curve is a tool for visualizing the performance and behavior of a machine learning model as the size of the training data increases. In the context of the Logistic Regression Forest in KAXAI, the Learning Curve depicts the model's performance on both the training and validation sets as a function of the number of training examples used.

By analyzing the Learning Curve, valuable insights can be gained into the behavior of the Logistic Regression Forest model during the learning process. For instance, the curve can reveal information about the bias-variance trade-off. With a small number of training examples, the model may exhibit high bias, resulting in low scores on both sets. As more training examples are used, the model's bias decreases, leading to improved scores. However, if overfitting occurs, the validation score may plateau or decrease, indicating an increase in variance.

The Learning Curve also allows for an assessment of the model's performance in terms of its ability to generalize to unseen data. Low scores on both sets suggest underfitting and may require a more complex model or additional features. Conversely, if the training score is high while the validation score is significantly lower, it indicates overfitting and may necessitate regularization techniques to improve generalization.

\begin{figure}[h!]
  \centering
  \includegraphics[width=0.48\textwidth]{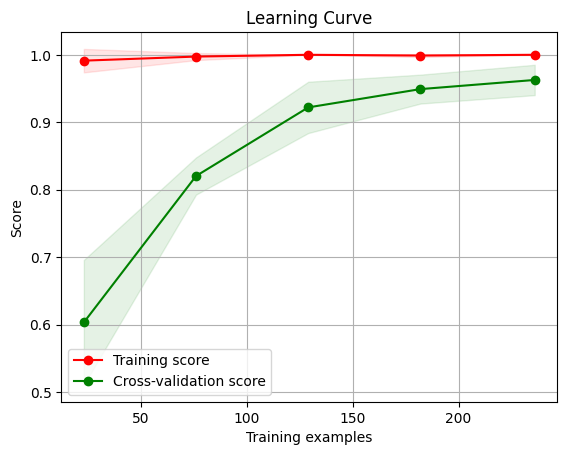}
  \hfill
  \includegraphics[width=0.48\textwidth]{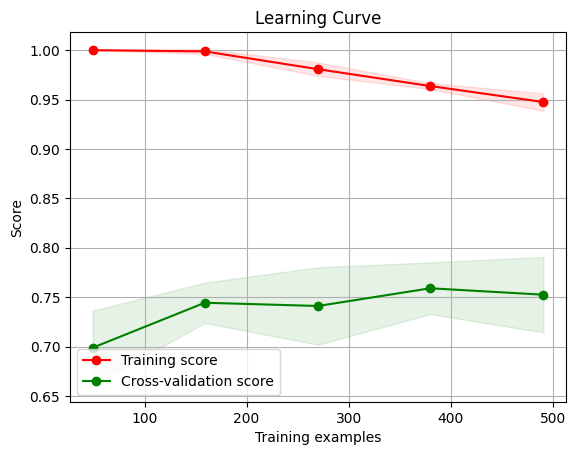}
  \caption{Learning Curve}
\end{figure}

\subsubsection{Precision Recall Curve}

The Precision-Recall Curve is a graphical representation that illustrates the trade-off between precision and recall for a classification model. In the case of the Logistic Regression Forest in KAXAI, the Precision-Recall Curve would provide insights into the model’s performance across different classification thresholds.

The curve enables a trade-off analysis between precision and recall. As the classification threshold decreases, both precision and recall typically decrease due to an increase in false positives. The curve displays the varying precision-recall pairs at different thresholds, allowing for the selection of an appropriate threshold based on the problem’s specific requirements.

The Precision-Recall Curve also facilitates model selection and comparison. Multiple models can be compared based on their curves, with a model whose curve is closer to the top-right corner generally performing better. Additionally, the curve aids in selecting an appropriate decision threshold based on the project’s objectives and constraints.

\begin{figure}[h!]
  \centering
  \includegraphics[width=0.48\textwidth]{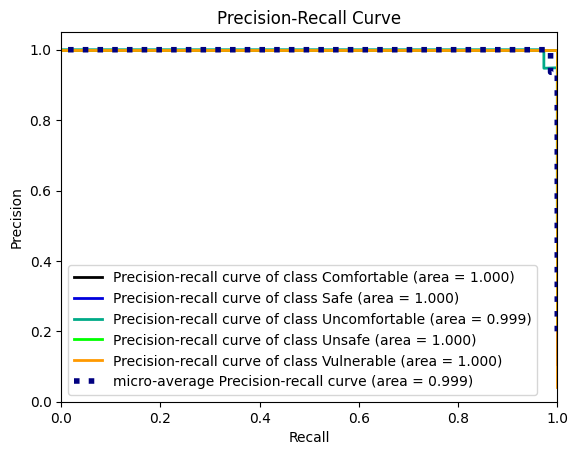}
  \hfill
  \includegraphics[width=0.48\textwidth]{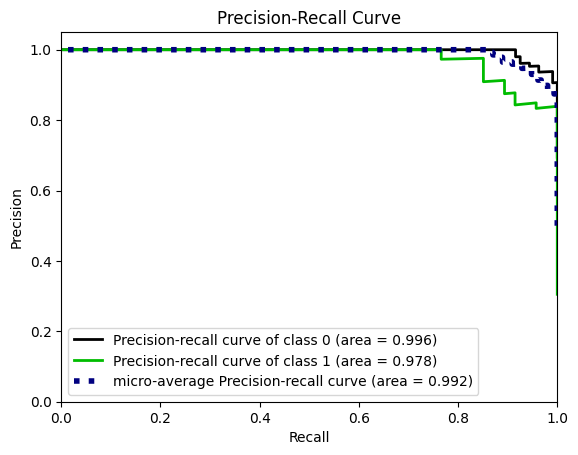}
  \caption{Precision Recall Curve}
\end{figure}

\subsubsection{ROC Curve}

The Receiver Operating Characteristic (ROC) curve is a graphical representation that displays the performance of a classification model, such as Logistic Regression Forest, at various classification thresholds. The curve provides insights into the model’s performance and allows for trade-off analysis between the True Positive Rate (TPR) and False Positive Rate (FPR). Additionally, the ROC curve can be used to compare the performance of different models and to select an appropriate classification threshold based on the specific requirements of the problem.

\begin{figure}[h!]
  \centering
  \includegraphics[width=0.48\textwidth]{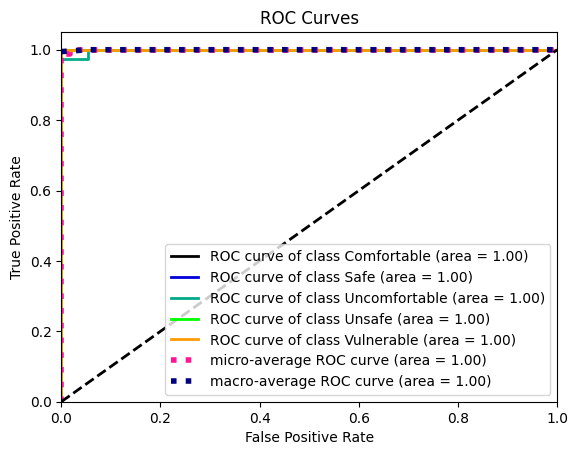}
  \hfill
  \includegraphics[width=0.48\textwidth]{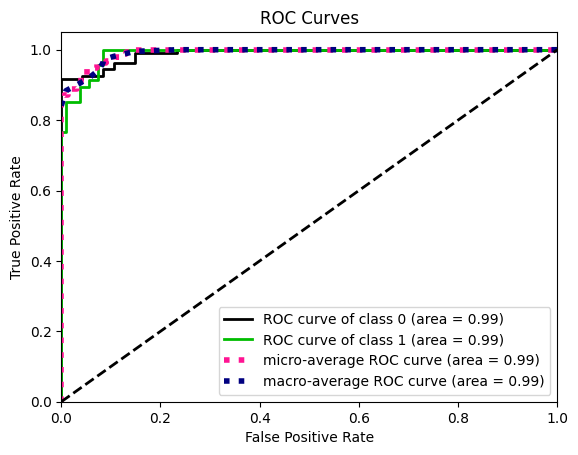}
  \caption{ROC Curve}
\end{figure}

\newpage
\subsubsection{Comparative Analysis of Logistic Regression Forest with Other Conventional Classifers}

The Logistic Regression Forest algorithm has demonstrated strong performance in comparison to other conventional classifiers, achieving high accuracy, precision, recall, and F1 score. With an accuracy of 96.73\%, it outperforms most classifiers except for Random Forest and CNN. Its precision, recall, and F1 score of 93.13\%, 92.15\%, and 93.38\%, respectively, indicate a robust overall predictive capability.

The algorithm’s high accuracy in the survey dataset suggests its ability to effectively capture underlying patterns and relationships in the data, resulting in more accurate predictions. This indicates its capacity to generalize well and make correct classifications on unseen data.

Moreover, the algorithm’s high recall in the diabetes dataset demonstrates its effectiveness in correctly identifying positive instances (diabetic cases) out of all actual positive cases. This is particularly significant in the context of medical diagnosis, as a high recall ensures that a greater proportion of diabetic cases are correctly identified, minimizing the chances of false negatives and providing valuable information for early detection and treatment.

The design significance of Logistic Regression Forest lies in its combination of logistic regression and decision forest techniques. By leveraging the strengths of both approaches, the algorithm can capture linear and nonlinear relationships in the data, handle categorical and numerical features, and handle both binary and multi-class classification tasks. This design flexibility allows Logistic Regression Forest to adapt to different types of datasets and achieve competitive performance in various domains.

\subsubsection{Comparative Analysis of Support Vector Tree with Other Conventional Classifers}

The Support Vector Tree algorithm has demonstrated competitive performance when compared to other conventional classifiers, achieving high accuracy, precision, recall, and F1 score. With an accuracy of 97.38\%, it exhibits a strong ability to correctly classify instances.

In terms of precision, recall, and F1 score, with values of 88.83\%, 87.86\%, and 88.63\% respectively, the Support Vector Tree algorithm performs well in accurately identifying positive instances and minimizing false positives and false negatives. These metrics indicate the algorithm's ability to correctly classify instances, which is crucial in various applications.

The algorithm's high accuracy in the survey dataset suggests its effectiveness in capturing underlying patterns and relationships in the data, enabling accurate predictions. This indicates its capacity to generalize well and make correct classifications on unseen data.

Moreover, the algorithm's high recall in the diabetes dataset highlights its effectiveness in correctly identifying positive instances (diabetic cases) out of all actual positive cases. This is particularly important in medical diagnosis, where the identification of true positive cases is crucial for accurate detection and treatment.

The design significance of the Support Vector Tree lies in its utilization of the support vector machine (SVM) algorithm, known for its effectiveness in handling complex datasets and nonlinear relationships. By combining SVM with decision tree-based techniques, the Support Vector Tree inherits SVM's boundary optimization power and decision trees' interpretability. This allows it to handle both linear and nonlinear relationships and provide meaningful insights into the classification process.

\begin{figure}[h!]
  \centering
  \includegraphics[width=0.48\textwidth]{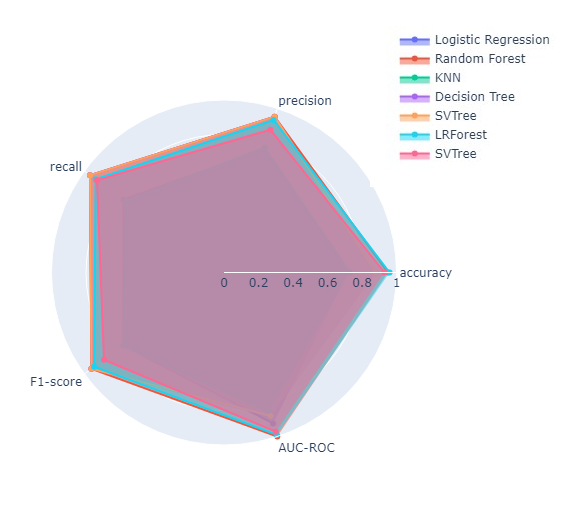}
  \hfill
  \includegraphics[width=0.48\textwidth]{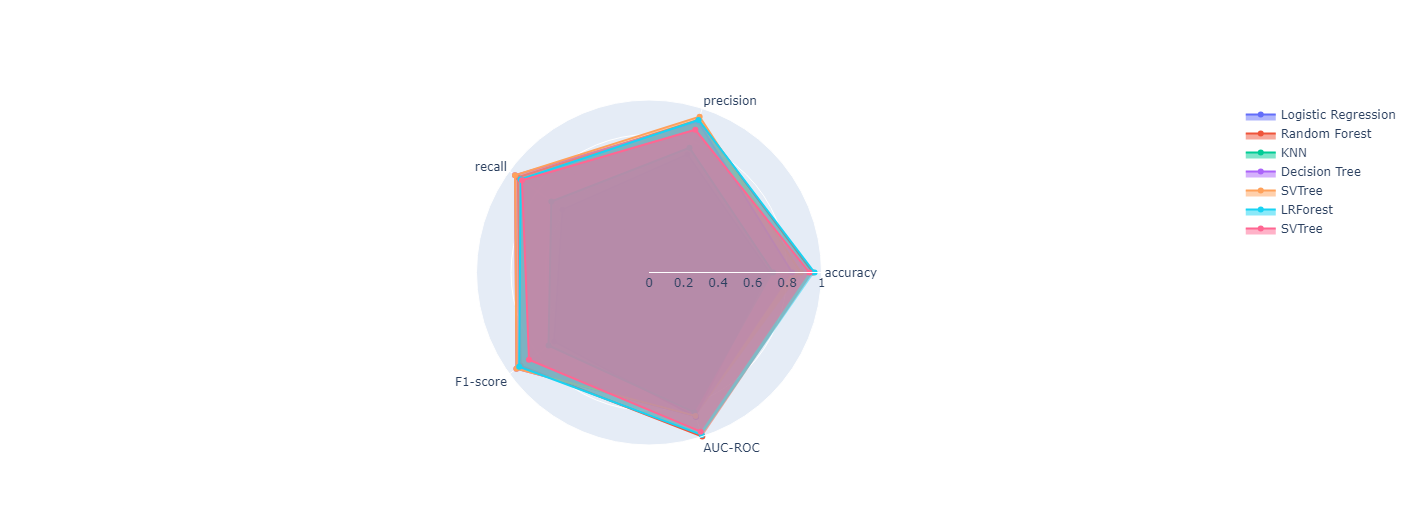}
  \caption{Spider Plot of Classification Report}
\end{figure}

\subsection{Interpretation Results of MEDLEY}

MEDLEY (Model Dependent Local Interpreter) is an interpretability technique that assigns preference scores to features for the predicted class in a classifier. These scores provide insights into the relevance and contribution of each feature to the classification decision. By analyzing these scores, we can understand which features are more influential in determining the predicted class.

The results obtained from MEDLEY can help us interpret and explain the decision-making process of the classifier. For example, if a specific feature receives a high preference score, it indicates that the classifier relied heavily on that feature to make its prediction. On the other hand, if a feature has a low preference score, it suggests that the classifier did not consider it as a crucial factor in its decision.

By examining these preference scores, we can infer the reasons behind the classifier’s decision. This interpretability can be valuable in various domains, such as healthcare, finance, and fraud detection, where understanding the reasons behind a classifier’s predictions is crucial for trust, accountability, and decision support.

It is important to note that MEDLEY is a model-dependent interpreter, meaning it relies on the specific classifier being analyzed. Different classifiers may have different preference scores for the same features, reflecting their unique decision-making processes. Therefore, the interpretability provided by MEDLEY is tailored to the particular classifier being interpreted.

\begin{figure}[h!]
  \centering
  \includegraphics[width=0.48\textwidth]{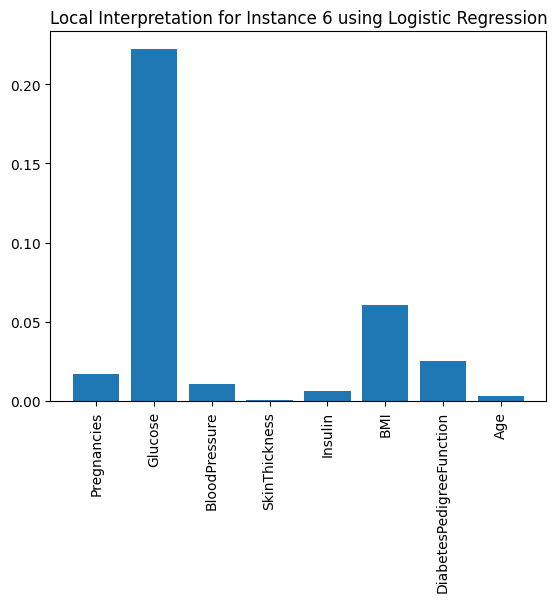}
  \hfill
  \includegraphics[width=0.48\textwidth]{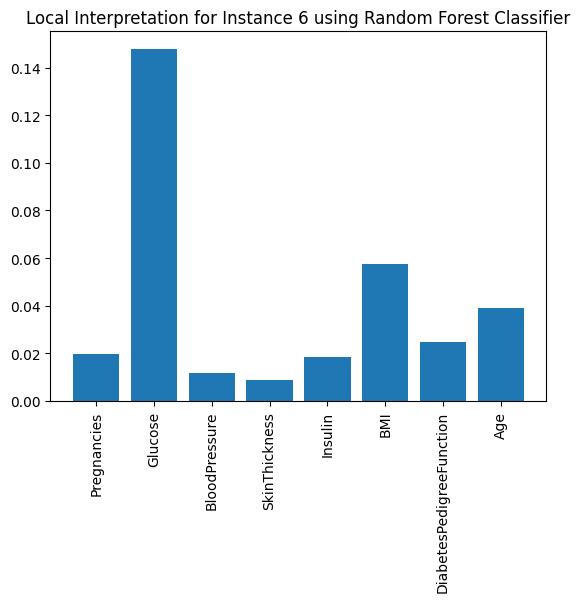}
  \caption{MEDLEY Interpretation of Sixth Instance of Diabetes Dataset}
\end{figure}

MEDLEY is a model-dependent interpreter, meaning it relies on the specific classifier being analyzed.

In the experiment using logistic regression and random forest classifiers, the magnitude of the preference scores changed between these two classifiers. However, the overall pattern for the dominant features remained unchanged. This demonstrates that while MEDLEY is sensitive to the classifier being used, it is also reliable in identifying the most influential features in determining the predicted class.

By analyzing these preference scores, we can understand which features are more influential in determining the predicted class. This information helps in interpreting and explaining the decision-making process of the classifier. The interpretability provided by MEDLEY can be valuable in various domains where understanding the reasons behind a classifier’s predictions is crucial for trust, accountability, and decision support.

\begin{figure}[h!]
  \centering
  \includegraphics[width=0.48\textwidth]{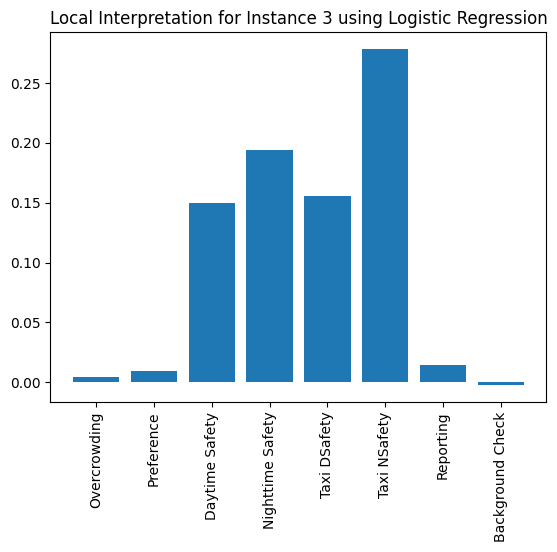}
  \hfill
  \includegraphics[width=0.48\textwidth]{Figures/MEDLEYRf.png}
  \caption{MEDLEY Interpretation of Sixth Instance of Diabetes Dataset}
\end{figure}

\subsubsection{Evaluating the Results of the MEDLEY}

First, for evaluating the MEDLEY interpretation, both permutation scores and local interpreter preference scores were plotted. Despite differences in magnitude, the pattern was the same, indicating that both methods identified the same dominant features as being influential in determining the predicted class.

Permutation importance is a technique that estimates feature importance by measuring the decrease in model performance when feature values are permuted. MEDLEY, on the other hand, assigns preference scores to features for the predicted class in a classifier, providing insights into the relevance and contribution of each feature to the classification decision.

Despite differences in magnitudes, permutation scores and local interpreter preference scores often exhibit similar patterns across features, highlighting those that have a significant impact on the classifier’s decision-making process. Features with higher permutation scores are likely to have higher preference scores in the local interpreter, indicating their importance in the classification process.

It is worth noting that while permutation scores provide a global measure of feature importance, local interpreter preference scores focus on the importance of a specific instance or class. Therefore, they can complement each other in providing a comprehensive understanding of feature relevance and the classifier’s decision process.

\begin{figure}[h!]
  \centering
  \includegraphics[width=0.58\textwidth]{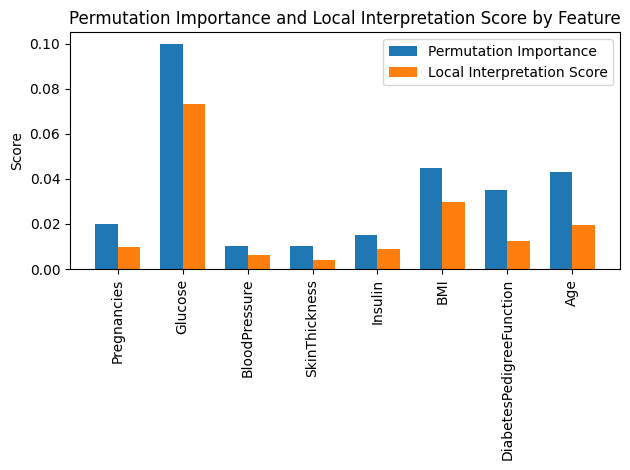}

  \caption{Permutation Importance, Local Interpretation Comparison}
\end{figure}

An experiment compared the performance of MEDLEY with LIME, Parzen, and Greedy by plotting the recall on truly important features. Results showed that MEDLEY performed better than Parzen and Greedy with a recall of 84\%, and slightly less than LIME which had a recall of 87\%.

Recall on truly important features measures the explainer's ability to correctly identify all relevant instances. In this case, it measures the ability of each interpretability technique to correctly identify the truly important features.

These results suggest that MEDLEY is a reliable interpretability technique that can accurately identify the most influential features in determining the predicted class, with performance comparable to other popular interpretability techniques.

Analyzing the preference scores assigned by MEDLEY allows us to understand which features are more influential in determining the predicted class. This information helps in interpreting and explaining the decision-making process of the classifier and can be valuable in various domains where understanding the reasons behind a classifier’s predictions is crucial for trust, accountability, and decision support.

\begin{figure}[h!]
  \centering
  \includegraphics[width=0.58\textwidth]{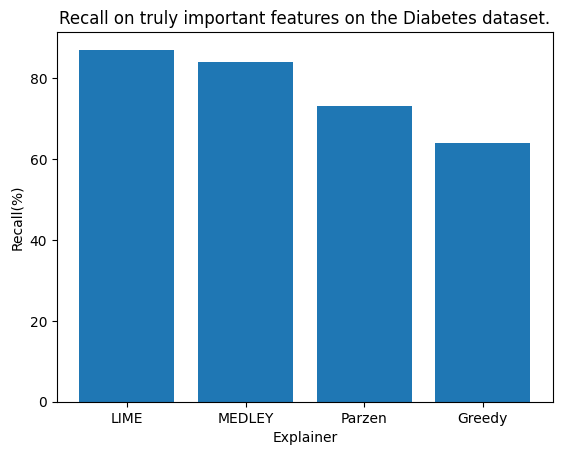}

  \caption{Comparative Recall on Truly Important features}
\end{figure}

In the experiment, we compared the performance of LIME and MEDLEY by plotting their results. Observations showed that both explainers chose the same features for explaining the classification decision and predicted the same pattern of feature importance, despite LIME being a model-agnostic explainer.

Dominant features had higher scores in both explainers, while weaker features had lower scores. This suggests that both LIME and MEDLEY are reliable interpretability techniques that can accurately identify the most influential features in determining the predicted class.

\begin{figure}[h!]
  \centering
  \includegraphics[width=0.48\textwidth]{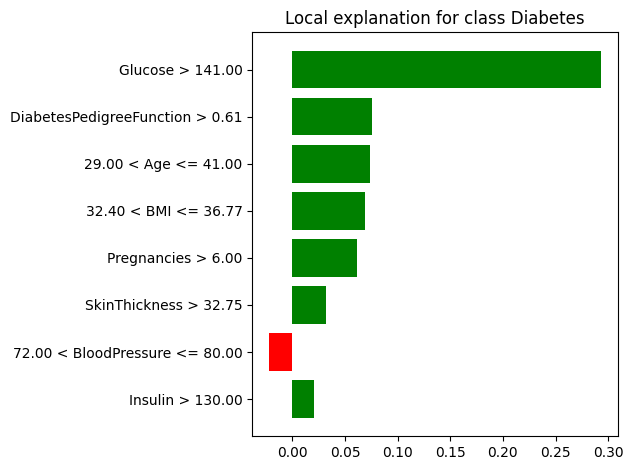}
  \hfill
  \includegraphics[width=0.48\textwidth]{Figures/MEDLEYRf.png}
  \caption{Comparative Analysis of LIME and MEDLEY}
\end{figure}

LIME explains the predictions of any classifier by approximating it locally with an interpretable model. MEDLEY (Model Dependent Local Interpreter), on the other hand, assigns preference scores to features for the predicted class in a classifier. The consistency in pattern between the two methods suggests their reliability in identifying the most influential features in determining the predicted class.

\subsection{Synthetic Data Generation with Large Language Model}

In KAXAI, the GPT-3.5 Turbo model is utilized for generating synthetic data. This advanced language model, developed by OpenAI, is designed to understand and generate human-like text based on the input it receives.

GPT-3.5 Turbo offers several advantages for synthetic data generation in KAXAI. It excels in generating text that closely resembles human language and can generate synthetic data that is coherent, fluent, and contextually appropriate. The model has been trained on a diverse range of texts across various domains, allowing it to generate synthetic data across a wide array of topics and industries.

GPT-3.5 Turbo can handle different input formats and can generate structured data, unstructured text, or even mimic specific data formats based on the provided instructions. It benefits from improvements in contextual understanding compared to earlier versions and can comprehend the context of the data being generated.

In response to a user prompt requesting the generation of a carbon emission dataset with 4 columns and 10 rows, GPT-3.5 Turbo, an advanced language model developed by OpenAI, would generate a synthetic dataset that meets the specified requirements. GPT-3.5 Turbo is capable of generating human-like text based on the input it receives. When provided with a prompt or query, the model generates a response that is coherent, fluent, and contextually appropriate.

In this instance, the user prompt specifies the desired structure and content of the synthetic dataset. GPT-3.5 Turbo would utilize its internal language model and contextual understanding to generate a dataset with 4 columns and 10 rows containing data related to carbon emissions.

\begin{table}[h!]
\centering
\caption{Synthetically Generated Carbon Emissions Dataset}
\begin{tabular}{lcccc}
\toprule
Country & Year & CO2 Emissions (kt) & CO2 Emissions per capita (metric tons) \\
\midrule
USA     & 2010 & 5,395,532          & 17.6                                   \\
China   & 2010 & 8,286,892          & 6.2                                    \\
India   & 2010 & 1,708,505          & 1.4                                    \\
Russia  & 2010 & 1,677,115          & 11.8                                   \\
Japan   & 2010 & 1,155,554          & 9.1                                    \\
Germany & 2010 & 798,565            & 9.9                                    \\
Canada  & 2010 & 541,020            & 15.9                                   \\
UK      & 2010 & 491,324            & 7.9                                    \\
Brazil  & 2010 & 422,598            & 2.2                                    \\
France  & 2010 & 365,666            & 5.6                                    \\
\bottomrule
\end{tabular}
\end{table}

In response to a user prompt requesting a dataset with 4 columns and 10 rows correlating happiness with GDP, GPT-3.5 Turbo generates a synthetic dataset meeting the specified requirements. 

While there are various techniques for evaluating the statistical properties of synthetic data, such as comparing the distribution and correlation of variables to those of real data, it is ultimately up to human evaluation to determine the significance and relevance of the generated data.

Human evaluation involves subjectively assessing the quality and usefulness of the synthetic data based on domain knowledge and expertise. This can include checking for accuracy, consistency, and plausibility of the data, as well as assessing its relevance to the research question or problem at hand.

It’s important to carefully evaluate synthetic data before using it for analysis or decision-making to ensure that it accurately reflects the desired characteristics and relationships. This can help to avoid erroneous conclusions or decisions based on inaccurate or misleading data.

\begin{table}[h!]
\centering
\caption{Synthetically Generated GDP and Happiness Dataset}
\begin{tabular}{lcccc}
\toprule
Country & Happiness Rank & Happiness Score & GDP per Capita \\
\midrule
Norway & 1 & 7.537 & 1.616463 \\
Denmark & 2 & 7.522 & 1.482383 \\
Iceland & 3 & 7.504 & 1.480633 \\
Switzerland & 4 & 7.494 & 1.56498 \\
Finland & 5 & 7.469 & 1.443572 \\
Netherlands & 6 & 7.377 & 1.503945 \\
Canada & 7 & 7.316 & 1.479204 \\
New Zealand & 8 & 7.314 & 1.405706 \\
Sweden & 9 & 7.284 & 1.494387 \\
Australia & 10 & 7.284 & 1.484415 \\
\bottomrule
\end{tabular}
\end{table}

\subsubsection{Synthetic Data Generation with Libraries }

In KAXAI, Python libraries were integrated to generate synthetic datasets for cryptocurrency and car data. However, the entries in these datasets appear to be randomly generated and therefore unreliable.

Several factors can contribute to the perceived unreliability of generating datasets with libraries. These include simplified models that may not fully reflect the complexity and diversity present in real datasets; limited training data that may not cover all possible variations and patterns present in real datasets; inherent bias or overgeneralization of patterns present in the training data; and a lack of domain-specific knowledge required to capture the unique characteristics and relationships within the data. These limitations can impact the ability of synthetic data generation libraries to generate realistic and representative synthetic datasets.

\begin{table}[h!]
\centering
\caption{Synthetically Generated Cryptocurrency Dataset}
\begin{tabular}{lcccc}
\toprule
Date & Crypto Name & Price (USD) & Market Cap (USD) \\
\midrule
2022-01-01 & Bitcoin & 45000 & 850000000000 \\
2022-01-02 & Ethereum & 3500 & 400000000000 \\
2022-01-03 & BinanceCoin & 500 & 80000000000 \\
2022-01-04 & Cardano & 2 & 70000000000 \\
2022-01-05 & Solana & 150 & 45000000000 \\
2022-01-06 & XRP & 1 & 40000000000 \\
2022-01-07 & Polkadot & 30 & 35000000000 \\
2022-01-08 & Dogecoin & 0.2 & 30000000000 \\
2022-01-09 & Avalanche & 100 & 20000000000 \\
2022-01-10 & Chainlink & 25 & 15000000000 \\
\bottomrule
\end{tabular}
\end{table}

If the generated data appears to be randomly generated and unreliable, it may indicate that the synthetic data generation process was not properly configured or that the chosen algorithm was not appropriate for the task at hand. In such cases, it may be necessary to re-evaluate the synthetic data generation process and make adjustments to improve the quality and reliability of the generated data.

\begin{table}[h!]
\centering
\caption{Synthetically Generated Car Dataset}
\begin{tabular}{lcccc}
\toprule
Car Make and Model & Year & Price (USD) & Mileage \\
\midrule
Toyota Camry & 2020 & 25000 & 10000 \\
Honda Civic & 2019 & 22000 & 15000 \\
Ford F-150 & 2018 & 30000 & 20000 \\
Chevrolet Silverado & 2017 & 35000 & 25000 \\
Nissan Rogue & 2016 & 27000 & 30000 \\
Jeep Wrangler & 2015 & 28000 & 35000 \\
Subaru Outback & 2014 & 26000 & 40000 \\
Hyundai Tucson & 2013 & 24000 & 45000 \\
Kia Sorento & 2012 & 29000 & 50000 \\
Mazda CX-5 & 2011 & 25000 & 55000 \\
\bottomrule
\end{tabular}
\end{table}

\subsubsection{Augmenting Orginal Dataset with GAN}

KAXAI’s ability to augment original datasets using Generative Adversarial Networks (GANs) and generate synthetic data is indeed considered a reliable way to enhance the dataset. GANs are a powerful technique in machine learning that involves training a generator network to produce synthetic samples that resemble the original data distribution.

One of the key reasons why GAN-based synthetic data generation is considered reliable is its ability to capture the underlying data distribution from the original dataset. By training the generator network to produce synthetic samples that closely resemble the original data, the augmented dataset can better capture the characteristics, patterns, and statistical properties of the real data.

In addition, the GAN-based augmentation process in KAXAI aims to preserve the integrity of the original dataset while adding synthetic samples. This means that the synthetic data is generated in a way that maintains the key features and relationships present in the original data, ensuring that the augmented dataset remains representative of the target domain.

The findings of your study are quite interesting. The fact that the feature importance calculated for both the original and synthetic datasets using the Random Forest Classifier showed remarkable similarity suggests that the synthetic data generated by KAXAI possesses validity and can be effectively used for training purposes.

The dominant features, namely Taxi NSafety, Taxi Dsafety, Nighttime Safety, and Daytime Safety, were identified as important in both datasets. This indicates that the model interpreted the synthetic data in a similar manner to the original data, emphasizing the importance of these features for classification.

The similarity in feature importance between the original and synthetic datasets is a positive outcome. It suggests that the synthetic data captures the underlying patterns and characteristics of the original data, enabling the model to learn and interpret them consistently. This finding supports the notion that synthetic data can serve as a viable alternative to original data in various applications.

\begin{figure}[h!]
  \centering
  \includegraphics[width=0.48\textwidth]{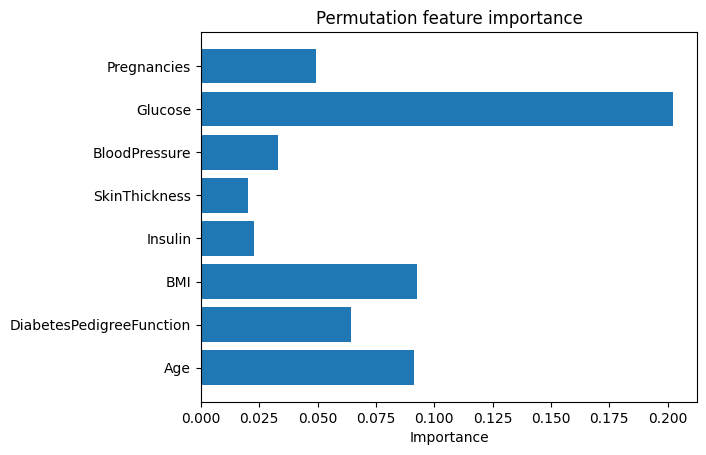}
  \hfill
  \includegraphics[width=0.48\textwidth]{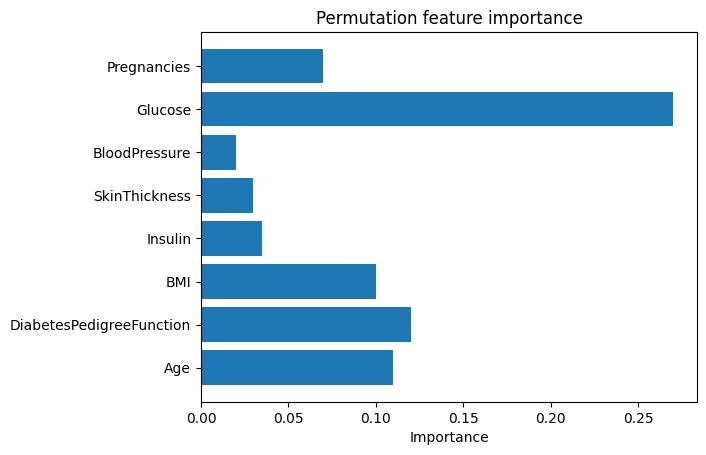}
  \caption{Feature Importance of Synthetic and Original Dataset}
\end{figure}

Measuring the standard deviation and applying the Kolmogorov-Smirnov (KS) test to compare the original and synthetic datasets are valuable steps in assessing the similarity and statistical properties of the two datasets.

The standard deviation is a measure of the dispersion or variability of data points around the mean. By calculating the standard deviation for both the original and synthetic datasets, one can evaluate how closely they resemble each other in terms of their spread. A smaller standard deviation indicates less variability and a higher degree of similarity between the datasets.

Applying the KS test allows for statistical comparison of the distributions of the two datasets. The KS test is a non-parametric test that assesses whether two datasets are drawn from the same underlying distribution. It compares the cumulative distribution functions (CDFs) of the datasets and calculates a test statistic that measures the maximum discrepancy between the two CDFs. The resulting p-value from the test indicates the level of similarity between the distributions.

\begin{figure}[h!]
  \centering
  \includegraphics[width=0.48\textwidth]{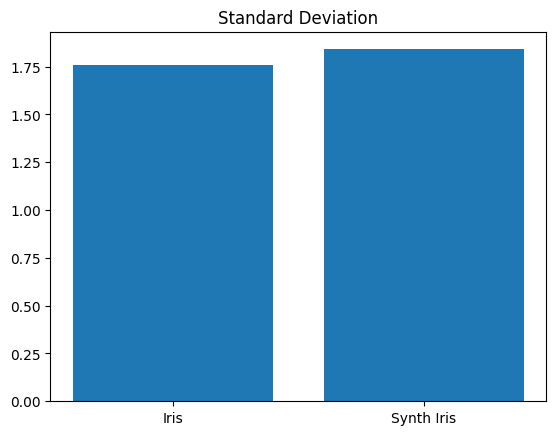}
  \hfill
  \includegraphics[width=0.48\textwidth]{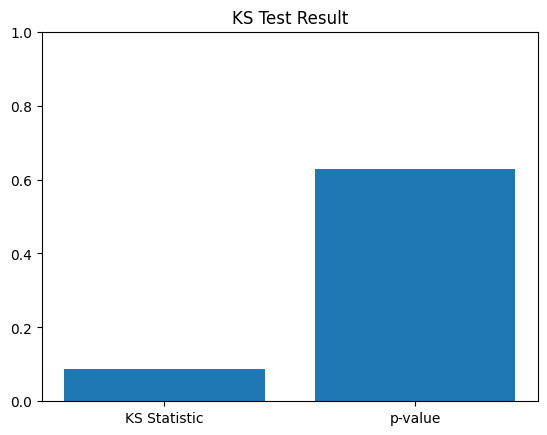}
  \caption{Standard Deviation and KS test of Synthetic and Original Dataset (Survey)}
\end{figure}

For fig-14, a KS statistic of 0.1 indicates that the maximum difference between the empirical CDFs of the two samples is 0.1. A p-value of 0.65 indicates that there is a 65\% probability of observing a KS statistic as extreme or more extreme than 0.1 if the two samples are drawn from the same distribution. Since this p-value is not small (i.e., it is greater than 0.05), we cannot reject the null hypothesis and conclude that there is insufficient evidence to suggest that the two samples are drawn from different distributions.

\begin{figure}[h!]
  \centering
  \includegraphics[width=0.48\textwidth]{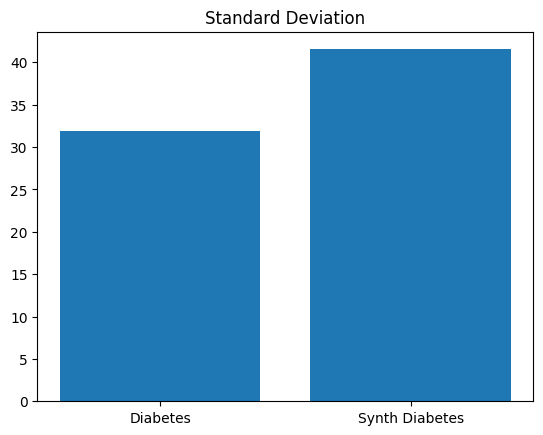}
  \hfill
  \includegraphics[width=0.48\textwidth]{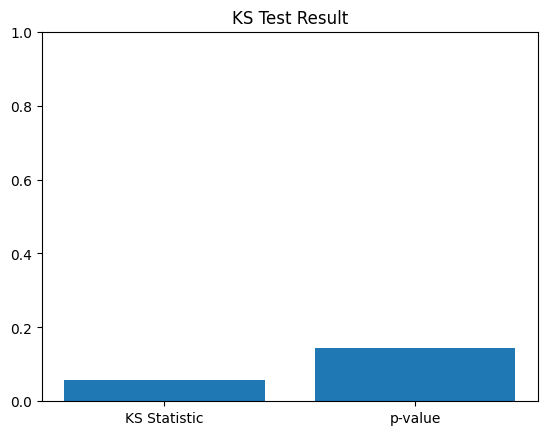}
  \caption{Standard Deviation and KS test of Synthetic and Original Dataset (Diabetes)}
\end{figure}
For fig-15, a KS statistic of 0.05 indicates that the maximum difference between the empirical CDFs of the two samples is 0.05. A p-value of 0.15 indicates that there is a 15\% probability of observing a KS statistic as extreme or more extreme than 0.05 if the two samples are drawn from the same distribution. Since this p-value is not small (i.e., it is greater than 0.05), we cannot reject the null hypothesis and conclude that there is insufficient evidence to suggest that the two samples are drawn from different distributions.

\begin{figure}[h!]
  \centering
  \includegraphics[width=0.48\textwidth]{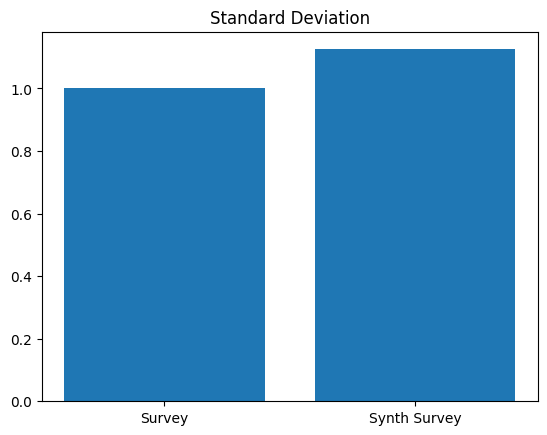}
  \hfill
  \includegraphics[width=0.48\textwidth]{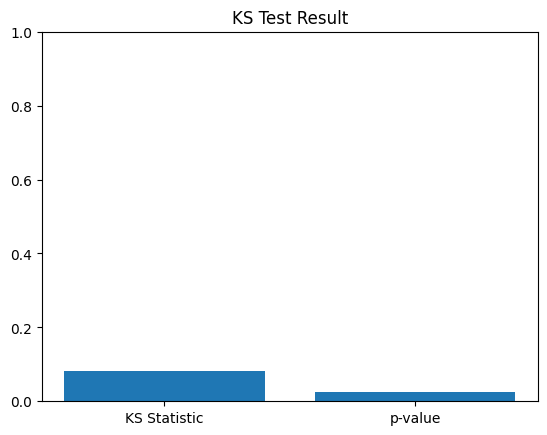}
  \caption{Standard Deviation and KS test of Synthetic and Original Dataset (Diabetes)}
\end{figure}

But in fig-16, we have to reject the null hypothesis because the KS statistics of 0.1 indicates the difference between the empirical CDFs of the two samples is 0.05, but the p-value is less than 0.05. When working with a categorical dataset, the values of each feature are discrete and represented using label encoding. Label encoding assigns numeric labels to each category in the categorical features, but these numeric labels do not carry the same meaning as continuous values. As a result, the continuous values of a feature are no longer present, making it difficult for GANs to accurately capture the underlying data distribution.

GANs are primarily designed to model continuous data distributions and generate samples from them. They typically operate on continuous latent spaces and generate samples by mapping points from this continuous space to the data space. In the case of categorical features, the discrete values lack the notion of continuity that allows GANs to effectively model the data distribution.

As a result, when using GANs for synthetic data generation on categorical datasets, the results may not be as successful as with continuous datasets. The GAN may struggle to generate realistic samples that accurately represent the complexity and variability of the original categorical dataset. The synthetic samples may not capture the nuances, variations, and dependencies present in the original categorical data.

\section{Discussion}

KAXAI’s outcomes align closely with its stated research goals. The system evaluates the effectiveness of synthetic data in approximating real data distributions using techniques such as GANs and LLM. KAXAI also explores the sensitivity of model outcomes to different classifiers, such as Logistic Regression, Random Forest, and CNN, providing insights into their strengths and weaknesses. Additionally, KAXAI assesses the user experience and effectiveness of ML tools integrated within the system, allowing users to interact with machine learning tools and evaluate their usability in practical scenarios. KAXAI introduces a model-dependent interpreter called MEDLEY, which provides insights into how a classifier makes decisions by assigning preference scores to each feature for the predicted class. Finally, KAXAI proposes and evaluates two novel classifiers, introducing new approaches and techniques to improve model performance and generalization. These contributions help advance the field of machine learning and provide valuable insights for practical applications.

\begin{table}[h!]
\caption{A comparative analysis with related works}
\begin{tabular}{p{1cm}p{3cm}p{2.5cm}p{3cm}p{3cm}p{2.5cm}}
\toprule
Title & Methodology & Limitations & Advantages & Contributions & Future Directions \\ 
\midrule
KAXAI & Data Flow-Oriented Design & Lack of Human Evaluation & Interactive platform. & Introduces a model-dependent interpreter called MEDLEY. Proposes and evaluates two novel classifiers. & Interactive Data Environment Design \\[1cm]
\cite{hall2009weka} & Data Flow-Oriented Design & Lack of Interpretability & Iterative Feature Selection & High Usuability & Not Specified \\[1cm]
\cite{he2021automl} & Comprehensive review of AutoML methods & Partial Opinion & Not Specified & Field Knowledge Popularization & Continuation of Exploring AutoML Fields \\[1cm]
\cite{olson2016tpot} & Data Structure-Oriented Design & Lack of Visualizations & Open source genetic programming-based AutoML system & Optimizes a series of feature preprocessors & Introducing Game theoretic Approach. \\[1cm]
\cite{jin2019auto} & Object Oriented Design & Lack of Data Preprocessing Feature  & Efficient neural architecture search system.  & Faster Implementation of Classfiers  & Not specified.  \\
\bottomrule
\end{tabular}
\end{table}

\begin{table}[h!]
\caption{Comparative Analysis with Other Conventional Classifiers}
\begin{tabular}{p{4cm}p{2cm}p{2cm}p{2cm}p{2cm}p{2cm}}
\toprule
Algorithm & Dataset & Accuracy & Precision & Recall & F1 Score \\
\midrule
Logistic Regression & Survey & 68.15\% & 48.79\% & 72.10\% & 48.97\% \\
 & Diabetes & 88.46\% & 88.50\% & 88.46\% & 88.42\% \\

K-Nearest Neighbor & Survey & 63.46\% & 60.77\% & 61.54\% & 59.72\% \\
 & Diabetes & 78.85\% & 79.62\% & 79.62\% & 75.60\% \\
Random Forest & Survey & 91.46\% & 91.67\% & 94.46\% & 94.83\% \\
 & Diabetes & 96.38\% & 89.17\% & 88.46\% & 88.67\% \\
Decision Tree & Survey & 59.62\% & 63.14\% & 61.54\% & 61.44\% \\
 & Diabetes & 76.92\% & 77.54\% & 75.00\% & 75.45\% \\
Support Vector Machine & Survey & 57.69\% & 58.17\% & 57.69\% & 57.82\% \\
 & Diabetes & 86.54\% & 91.03\% & 86.54\% & 86.53\% \\

Naive Bayes  & Survey & 55.77\% & 41.07\% & 55.77\% & 47.23\% \\
 & Diabetes & 73.08\% & 73.46\% & 73.08\% & 70.47\% \\

ZeroR & Survey & 26.92\% & 07.24\% & 26.92\% & 11.42\% \\
 & Diabetes  & 26.92\% & 07.24\% & 26.92\% & 11.42\% \\

CNN & Survey & 92.77\% & 91.08\% & 89.63\% & 90.72\% \\
 & Diabetes  & 95.31\% & 93.33\% & 95.82\% & 92.64\% \\

\textbf{LRForest} & Survey & \textbf{96.73\%} & 93.13\% & 92.15\% & 93.38\% \\
 & Diabetes  & 88.75\% & 92.23\% & \textbf{93.86\%} & 92.64\% \\

\textbf{SVTree} & Survey & 93.15\% & 88.08\% & 88.63\% & 87.72\% \\
 & Diabetes  & \textbf{97.38\%} & 88.83\% & 87.86\% & 88.63\% \\

\bottomrule
\end{tabular}
\end{table}

\begin{figure}[h!]
  \centering
  \includegraphics[width=0.48\textwidth]{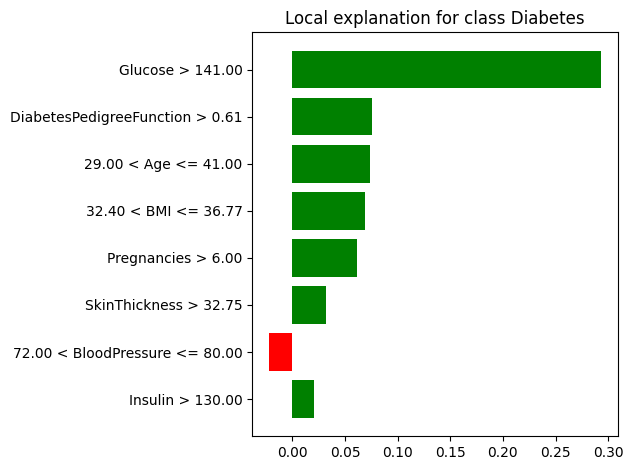}
  \hfill
  \includegraphics[width=0.48\textwidth]{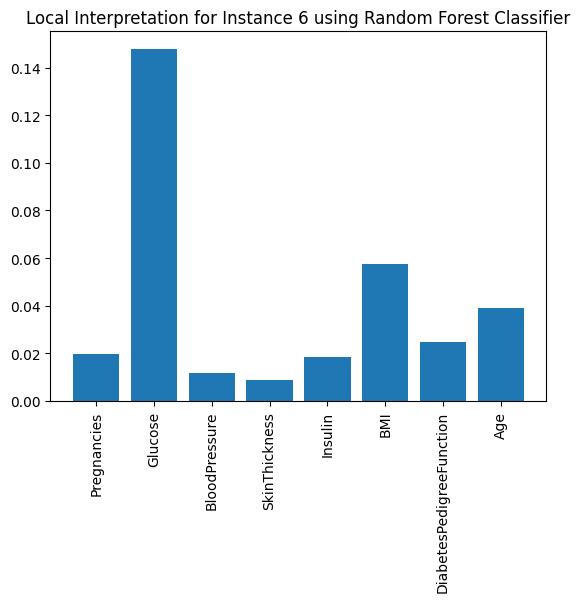}
  \caption{Comparative Analysis of LIME and MEDLEY}
\end{figure}

\newpage
\section{Limitations}

KAXAI evaluates synthetic data, explores model sensitivity, assesses ML tools, introduces MEDLEY, and proposes novel classifiers, advancing the field of machine learning. Nonetheless, the work has encountered several obstacles:

\begin{itemize}
    \item The AutoML systems were not thoroughly tested, which may slightly impact the accuracy and reliability of the results.
    \item MEDLEY, the model-dependent interpreter introduced by KAXAI, has limited human evaluation making the evaluation enigmatic
    \item The experiments were conducted on a limited number of datasets thus generalization was difficult
\end{itemize}

\section{Conclusion}

The Research has successfully integrated AutoML, XAI, and synthetic data generation within a single platform. The system proposes and evaluates two novel classifiers, presenting new approaches and techniques to enhance model performance and generalization. KAXAI also evaluates the user experience and effectiveness of ML tools integrated within the system, enabling users to interact with machine learning tools and assess their practicality. The system introduces a model-dependent interpreter called MEDLEY, which offers insights into classifier decision-making by assigning preference scores to each feature for the predicted class. KAXAI investigates the sensitivity of model outcomes to different classifiers, including Logistic Regression, Random Forest, and CNN, providing insights into their strengths and limitations. Lastly, the system assesses the effectiveness of synthetic data in approximating real data distributions using techniques such as GANs and LLM. These contributions advance the field of machine learning and offer valuable insights for practical applications.

In the future, KAXAI could broaden its scope beyond tabular data and incorporate more reliable synthetic data generation techniques with Langchain and AutoGPT. The Model Interpreter can explain the synthetically generate output by incorporating locating and editing factual associations of LLM. Additionally, we want to develop an interactive user interface where users can communicate with their dataset using natural language.

\newpage
\bibliographystyle{unsrt}  
\bibliography{biblio}

\end{document}